\documentclass{article}

\usepackage[final,nonatbib]{nips_2018}

\usepackage[utf8]{inputenc} 
\usepackage[T1]{fontenc}    
\usepackage{url}            
\usepackage{booktabs}       
\usepackage{amsfonts}       
\usepackage{nicefrac}       
\usepackage{microtype}      
\usepackage{array}			

\usepackage{bm}
\usepackage{algorithm}
\usepackage[noend]{algorithmic}
\usepackage{amsmath}
\usepackage{amssymb}
\usepackage{graphicx}
\usepackage{caption}
\usepackage{subcaption}
\usepackage{epstopdf}
\usepackage{wrapfig}
\usepackage{helvet}
\usepackage{courier}
\usepackage{epsfig}
\usepackage{amsfonts}
\usepackage{amsthm}

\usepackage[numbers,sort&compress]{natbib}
\usepackage{natbib}

\newtheorem{theorem}{Theorem}

\newtheorem{lemma}{Lemma}

\newcommand{\squishlisttwo}
{\begin{list}{$\bullet$}{ 
			\setlength{\itemsep}{3pt}
			\setlength{\parsep}{0pt}
			\setlength{\topsep}{0pt}
			\setlength{\partopsep}{0pt}
			\setlength{\leftmargin}{1em}
			\setlength{\labelwidth}{1.5em}
			\setlength{\labelsep}{0.5em}}}	
\newcommand{\squishend}{\end{list}}

\title{Collective Online Learning via Decentralized Gaussian Processes in Massive Multi-Agent Systems}

\author{
	Trong Nghia Hoang \& Jonathan How\\
	Laboratory for Information and Decision Systems\\
	Massachusetts Institute of Technology\\
	\texttt{{nghiaht,jhow}@mit.edu} \\
	\And
	Quang Minh Hoang \& Kian Hsiang Low\\
	School of Computing\\
	National University of Singapore\\
	\texttt{{hqminh,lowkh}@comp.nus.edu.sg} 
}

\begin{document}
\newcounter{sol} 
\setcounter{sol}{1} 

\maketitle
\begin{abstract}
\vspace{-2mm}
Distributed machine learning (ML) is a modern computation paradigm that divides its workload into independent tasks that can be simultaneously achieved by multiple machines (i.e., agents) for better scalability. However, a typical distributed system is usually implemented with a central server that collects data statistics from multiple independent machines operating on different subsets of data to build a global analytic model. This centralized communication architecture however exposes a single choke point for operational failure and places severe bottlenecks on the server's communication and computation capacities as it has to process a growing volume of communication from a crowd of learning agents. To mitigate these bottlenecks, this paper introduces a novel \underline{Co}llective \underline{O}nline \underline{L}earning \underline{G}aussian \underline{P}rocess (COOL-GP) framework for massive distributed systems that allows each agent to build its local model, which can be exchanged and combined efficiently with others via peer-to-peer communication to converge on a global model of higher quality. Finally, our empirical results consistently demonstrate the efficiency of our framework on both synthetic and real-world datasets.\vspace{-2mm} 
\end{abstract}

\section{Introduction}
\label{intro}
%
Distributed \emph{Gaussian process} (GP) models \cite{LowUAI13,Marc15,Yarin14,NghiaICML16,LowAAAI15} are conventionally designed with a server-client paradigm where a server distributes the computational load among parallel machines (i.e., client nodes) to achieve scalability to massive, streaming datasets.
This paradigm can potentially allow the richness and expressive power of GP models \cite{Rasmussen06} (Section~\ref{fgp}) to be exploited by multiple mobile sensing agents for distributed inference of the complex latent behavior and correlation structure underlying their local data. 
Such a prospect has inspired the recent development of distributed GP fusion algorithms \cite{LowUAI12,LowRSS13,Arik15,Rakshit17}: Essentially, the ``client'' agents encapsulate their own local data into memory-efficient local summary statistics based on a \emph{common} set of \emph{fixed/known} GP hyperparameters and \emph{inducing inputs}, then communicate them to some ``server'' agent(s) to be fused into globally consistent summary statistics. These will in turn be sent back to the ``clients'' for predictive inference.

These distributed GP fusion algorithms inherit the advantage of being adjustably lightweight
by restricting the number of inducing inputs (hence the size of the local and global summary statistics) to fit the agents' limited computational and communication capabilities at the expense of predictive accuracy. However, such algorithms fall short of achieving the truly decentralized GP fusion necessary for scaling up to a massive number of agents grounded in the real world (e.g., traffic sensing, modeling, and prediction by autonomous vehicles cruising in urban road networks \cite{Arik15,NghiaICML14,min11,wang05,Bayen10a}, distributed inference on a network of IoT and mobile devices \cite{Kang16,Sarkar14}) due to several critical issues. These includes:
(a) an obvious limitation is the single point(s) of failure with the server agent(s) whose computational and communication capabilities must be superior and robust;
(b) different mobile sensing agents are likely to gather local data of varying behaviors and correlation structure from possibly separate localities of the input space (e.g., spatiotemporal) and could therefore incur considerable information loss due to summarization based on a common set of fixed/known GP hyperparameters and inducing inputs, especially when the inducing inputs are few and far from the data (in the correlation sense); and 
(c) like their non-fusion counterparts, distributed GP fusion algorithms implicitly assume a one-time processing of a fixed set of data and would hence repeat the entire fusion process involving all local data gathered by the agents whenever new batches of streaming data arrives, which is potentially very expensive. Further problems could occur in the event of a transmission loss between the clients and server, which can happen when the locations of clients are changing over time (e.g., autonomous vehicles cruising an urban road network to collect traffic data \cite{Arik15}). This loss might prevent the prediction model from being generated \cite{LowUAI13} or as shown in Section~\ref{exp}, cause its performance to degrade badly due to irrecoverable loss.

To overcome these limitations, this paper presents a \emph{\underline{Co}llective \underline{O}nline \underline{L}earning via \underline{GP}} (COOL-GP) framework that enables a massive number of  agents to perform decentralized online GP fusion based on their own possibly different sets of \emph{learned} GP hyperparameters and inducing inputs.
A key technical challenge here lies in how the summary statistics currently being maintained by an agent can be fused efficiently in constant time and space with the summary statistics of a new batch of data or  another agent based on a possibly different set of GP hyperparameters and inducing inputs.
To realize this, we exploit the notion of a latent encoding vocabulary \cite{Candela05,Snelson07a,Titsias09,Miguel10,Hensman13,NghiaICML15,NghiaAAAI17} as a shared medium to exchange and fuse summary statistics of different batches of data or agents based on different sets of GP hyperparameters and inducing inputs (Section~\ref{sgps}).
This consequently enables us to design and develop a novel sampling scheme for efficient approximate online GP inference, a novel pairwise operator for fusing the summary statistics of different agents, and a novel decentralized message passing algorithm that can exploit sparse connectivity among agents for improving efficiency and enhance the robustness of our framework to transmission loss (Section~\ref{fusion}).
We provide a rigorous analysis of the approximation loss arising from the online update and fusion  
in Section~\ref{analysis}. Finally, we empirically evaluate the performance of COOL-GP on an extensive benchmark comprising both synthetic and real-world datasets with thousands of agents (Section~\ref{exp}).

\vspace{-4mm}
\section{Background and Notation}
\vspace{-2mm}
\label{fgp}
GP \cite{Rasmussen06} is a state-of-the-art model for predictive analytics due to its capacity to represent complex behaviors of data in highly sophisticated domains. Specifically, let $\mathbb{X} \subseteq \mathbb{R}^d$ represents an input domain and $\mathrm{f}: \mathbb{X} \rightarrow \mathbb{R}$ denotes a random function mapping each $d$-dimensional input feature vector $\mathbf{x} \in \mathbb{X}$ to a stochastic scalar measurement $\mathrm{f}(\mathbf{x}) \in \mathbb{R}$ and its noisy observation $\mathrm{y} \triangleq \mathrm{f}(\mathbf{x}) + \epsilon$ where $\epsilon \sim \mathcal{N}(0, \sigma^2)$. To characterize the stochastic behavior of $\mathrm{f}(\mathbf{x})$, a GP model assumes that for every finite subset of inputs $\mathbf{X}_\mathcal{D} \triangleq \{\mathbf{x}_1, \ldots, \mathbf{x}_n\} \subseteq \mathbb{X}$, the corresponding column vector $\mathbf{f}_\mathcal{D} \triangleq [\mathrm{f}(\mathbf{x}_1) \ldots \mathrm{f}(\mathbf{x}_n)]^\top$ of stochastic scalar measurements is distributed \emph{a priori} by a multivariate Gaussian distribution with mean $\mathbf{m}_\mathcal{D} \triangleq [\mathrm{m}(\mathbf{x}_1) \ldots \mathrm{m}(\mathbf{x}_n)]^\top$ and covariance $\mathbf{K}_\mathcal{DD} \triangleq [\mathrm{k}(\mathbf{x}_i,\mathbf{x}_j)]_{ij}$ induced from a pair of user-specified mean and covariance functions, $\mathrm{m}: \mathbb{X} \rightarrow \mathbb{R}$ and $\mathrm{k}: \mathbb{X} \times \mathbb{X} \rightarrow \mathbb{R}$, respectively. For notational simplicity, we assume a zero mean function $\mathrm{m}(\mathbf{x}) = 0$. Then, let $\mathbf{y}_\mathcal{D} \triangleq [\mathrm{y}_1 \ldots \mathrm{y}_n]^\top$ denotes the corresponding vector of noisy observations $\{\mathrm{y}_i\}_{i=1}^n$ where $\mathrm{y}_i \triangleq \mathrm{f}(\mathbf{x}_i) + \epsilon$ with $\epsilon \sim \mathcal{N}(0, \sigma^2)$, the posterior distribution over $\mathrm{f}(\mathbf{x}_\ast)$ for any test input $\mathbf{x}_\ast$ is Gaussian with mean $\mu(\mathbf{x}_\ast) = \mathbf{k}_\ast^\top(\mathbf{K}_\mathcal{DD} + \sigma^2 \mathbf{I})^{-1}\mathbf{y}_\mathcal{D}$ and variance $\sigma^2(\mathbf{x}_\ast) = \mathrm{k}(\mathbf{x}_\ast,\mathbf{x}_\ast) - \mathbf{k}_\ast^\top(\mathbf{K}_\mathcal{DD} + \sigma^2 \mathbf{I})^{-1}\mathbf{k}_\ast$ where $\mathbf{k}_\ast \triangleq [\mathrm{k}(\mathbf{x}_\ast,\mathbf{x}_1) \ldots \mathrm{k}(\mathbf{x}_\ast, \mathbf{x}_n)]^\top$. A complete predictive map over the (possibly infinite) input domain $\mathbb{X}$ can then be succinctly represented with $\{(\mathbf{K}_\mathcal{DD} + \sigma^2 \mathbf{I})^{-1}\mathbf{y}_\mathcal{D}, (\mathbf{K}_\mathcal{DD} + \sigma^2 \mathbf{I})^{-1}\}$.

This representation is not efficient because its size (computation) grow quadratically (cubically) in the size of data. More importantly, since the GP representation is specific to a particular data variation scale (i.e., the kernel parameters or hyper-parameters), it cannot be used as a common ground to facilitate communication between agents operating in related domains with different variation scales. To mitigate these issues, we instead represent each agent's local model using a common unit-scale GP and a transformation operator that warps the unit-scale GP into a domain-specific GP parameterized with different scale reflecting the variation in local data. Intuitively, this allows each agent to translate the statistical properties of its specific domain to those of a common domain and facilitates efficient communication between agents (Section~\ref{fusion}) while maintaining its own set of hyper-parameters.


Let $\mathrm{u}(\mathbf{z}) \sim \mathcal{GP} (0, \mathrm{k}_{\mathrm{uu}}(\mathbf{z},\mathbf{z}'))$ with
$\mathrm{k}_{\mathrm{uu}}(\mathbf{z},\mathbf{z'}) = \mathrm{exp}\left(-0.5(\mathbf{z} - \mathbf{z'})^\top(\mathbf{z} - \mathbf{z'})\right)$.
We can then characterize the distribution of a domain-specific function $\mathrm{f}(\mathbf{x})$ in terms of $\mathrm{u}(\mathbf{z})$ and its prior distribution $\mathcal{GP} (0, \mathrm{k}_{\mathrm{uu}}(\mathbf{z},\mathbf{z}'))$ over the unit-scale domain, which will be referred to as the standardized domain hereafter for convenience. In particular, let $\mathbf{W}$ be a projection matrix that maps domain-specific inputs $\mathbf{x} \in \mathbb{X}$ onto the standardized domain of $\mathbf{z}$ and the latent function $\mathrm{f}$ can be characterized in terms of $\mathrm{u}$ as $\mathrm{f}(\mathbf{x}) = \sigma_s \mathrm{u}(\mathbf{Wx})$. This implies $\mathrm{f}(\mathbf{x}) \sim \mathcal{GP}(0, \mathrm{k}_\mathrm{ff}(\mathbf{x,x'}))$ where \cite{Titsias13}
\begin{eqnarray}
\hspace{-1mm}\mathrm{k}_\mathrm{ff}(\mathbf{x,x'}) \triangleq \sigma_s^2\mathrm{exp}\left(-0.5(\mathbf{x - x'})^\top\mathbf{W}^\top\mathbf{W}(\mathbf{x - x'})\right) .
\label{eq:3.3}
\end{eqnarray}
Furthermore, it can be shown that the cross-domain covariance between $\mathrm{f}(\mathbf{x})$ and $\mathrm{u}(\mathbf{z})$ is also analytically tractable: $\mathrm{k}_{\mathrm{fu}}(\mathbf{x,z}) = \sigma_s \mathrm{exp} \left(-0.5(\mathbf{Wx - z})^\top(\mathbf{Wx - z})\right)$. This enables an inference of statistical properties of $\mathrm{u}(\mathbf{z})$ using observations of the domain-specific function $\mathrm{f}(\mathbf{x})$ via learning an appropriate projection matrix $\mathbf{W}$ (as detailed in the remaining of this section), which forms the basis for an efficient agent representation (Section~\ref{sgps}) amenable to cross-domain communication via the common function $\mathrm{u}(\mathbf{z})$ (Section~\ref{fusion}). 

The cost-efficient GP representation of a learning agent can be achieved via exploiting the vector $\mathbf{u} = [\mathrm{u}(\mathbf{z}_1) \ldots \mathrm{u}(\mathbf{z}_m)]^\top$ of latent inducing output or encoding vocabulary for a small set of $m$ standardized inputs  $\mathbf{Z} = \{\mathbf{z}_1, \ldots, \mathbf{z}_m\}$  to construct sufficient statistics for $\mathbf{y}_\mathcal{D}$. That is, for every test input $\mathbf{x}_\ast$ and $\mathbf{f}_\ast = \mathrm{f}(\mathbf{x}_\ast)$, we can characterize the predictive distribution $\mathrm{p}(\mathbf{f}_\ast | \mathbf{y}_\mathcal{D})$ in terms of the posterior $\mathrm{p}(\mathbf{u},\mathbf{W} |\mathbf{y}_\mathcal{D})$ which, in turn, induces a cost-efficient surrogate representation $\mathrm{q}(\mathbf{u}, \mathbf{W})$. This can be achieved by minimizing the KL-divergence between  $\mathrm{q}(\mathbf{f}_\mathcal{D},\mathbf{u},\mathbf{W}) \triangleq \mathrm{q}(\mathbf{u},\mathbf{W}) \mathrm{p}(\mathbf{f}_\mathcal{D}|\mathbf{u,W})$ and $\mathrm{p}(\mathbf{f}_\mathcal{D}, \mathbf{u}, \mathbf{W} | \mathbf{y}_\mathcal{D})$, which is equivalent to maximizing $\mathrm{L}(\mathrm{q}) \triangleq \mathbb{E}_{\mathrm{q}} \left[ \mathrm{log}\ \mathrm{p}(\mathbf{y}_\mathcal{D}|\mathbf{f}_\mathcal{D})\right] -  \mathrm{D_{KL}}(\mathrm{q}(\mathbf{u},\mathbf{W}) \| \mathrm{p}(\mathbf{u},\mathbf{W}))$. By parameterizing the prior $\mathrm{p}(\mathbf{u,W}) = \mathrm{p}(\mathbf{u})\mathrm{p}(\mathbf{W})$ where $\mathrm{p}(\mathbf{u}) \triangleq \mathcal{N}(\mathbf{u} | 0, \mathbf{K}_\mathcal{UU})$ with $\mathbf{K}_\mathcal{UU} \triangleq [\mathrm{k}_{\mathrm{uu}}(\mathbf{z}_i, \mathbf{z}_j)]_{i,j}$ and $\mathrm{p}(\mathbf{W})$ is a product of standard normals, it follows that the optimal marginal distribution $\mathrm{q}(\mathbf{W}) = \prod_{i=1}^d\prod_{j=1}^d\mathcal{N}(\mathrm{w}_{ij} | \mu_{ij}, \sigma^2_{ij})$. The agent's unique defining hyperparameters $\theta = \{\mu_{ij},\sigma_{ij}\}_{i,j}$ can then be optimized via gradient ascent of $\mathrm{L}(\mathrm{q})$, hence accounting for the data variation scale at its specific location.  Then, given $\mathrm{q}(\mathbf{W})$, $\mathrm{q}(\mathbf{u})$ is also a Gaussian whose mean $\mathbf{m}$ and covariance $\mathbf{S}$ can be analytically derived as\vspace{1mm}
\begin{eqnarray}
\hspace{-10mm}\mathbf{S} \ =\  \sigma^2_n\mathbf{K}_\mathcal{UU}(\sigma^2_n\mathbf{K}_\mathcal{UU} + \mathbf{C}_\mathcal{UU})^{-1}\mathbf{K}_\mathcal{UU} \ \ \ \ \text{;}\ \ \ \ 
\mathbf{m} \ =\  \mathbf{K}_\mathcal{UU}(\sigma^2_n\mathbf{K}_\mathcal{UU} + \mathbf{C}_\mathcal{UU})^{-1} \mathbf{C}_\mathcal{UD}\mathbf{y}_\mathcal{D}\vspace{1mm}
\label{eq:2.4}
\end{eqnarray}
where $\mathbf{K}_\mathcal{DU} \triangleq \left[\mathrm{k}_\mathrm{fu}(\mathbf{x}_i,\mathbf{z}_j)\right]_{i,j}, \mathbf{K}_\mathcal{UD} \triangleq \mathbf{K}_\mathcal{DU}^\top, \mathbf{C}_\mathcal{UU} \triangleq \mathbb{E}_{q(\mathbf{W})}\left[\mathbf{K}_\mathcal{UD}\mathbf{K}_\mathcal{DU} \right]$, and
$\mathbf{C}_\mathcal{UD} \triangleq \mathbb{E}_{q(\mathbf{W})}\left[\mathbf{K}_\mathcal{UD}\right]$. Eq.~\eqref{eq:2.4} yields an efficient representation $\{\mathbf{S},\mathbf{m},\theta \}$ of the posterior distribution $\mathrm{p}(\mathbf{u}, \mathbf{W} | \mathbf{y}_\mathcal{D}) \simeq \mathrm{q}(\mathbf{u})\mathrm{q}(\mathbf{W})$ which incurs linear computation and representation costs in the size of data. This enables the development of a communicable agent representation that can be updated efficiently when new data arrives and is amenable to cross-domain model fusion (Sections~\ref{sgps} and~\ref{pairwise}).

{\bf Remark 1.} The standardized inputs $\mathbf{Z}$ can be selected and optimized offline via simulation: different sets of synthetic data can be generated from the standardized domain and we select $\mathbf{Z}$ that yields the best averaged RMSE on those synthetic datasets (to ensure that $\mathbf{Z}$ best represents the domain). 
\vspace{-3mm}
\section{Agent Representation}
\vspace{-1mm}
\label{sgps}
Recomputation of the approximate posterior $\mathrm{q}(\mathbf{u})$ as new data arrives is often prohibitively expensive. This section presents a reparameterization of Eq.~\eqref{eq:2.4} achieved by exploiting the natural representation of $\mathrm{q}(\mathbf{u})$ that enables an efficient update of the reformulated parameters as new data arrives. We then show that the hyperparameters $\theta$ can also be learned online (Section~\ref{qw}) as an important extension of the prior decentralized ML literature, which assumes knowledge of hyperparameters \cite{Rakshit17,Arik15}.

\subsection{Online Update for Inducing Output Posterior}
\label{qu}
Let $\mathbf{R} = [\mathbf{R_1}; \mathbf{R}_2] \triangleq [\mathbf{S}^{-1}; \  \mathbf{S}^{-1}\mathbf{m}]$ denote the natural parameters of $\mathrm{q}(\mathbf{u})$. Eq.~\eqref{eq:2.4} can then be reparameterized in terms of $\mathbf{R}$ to reveal an additive decomposition across different blocks of data. That is, let $\{\mathcal{D}_1, \mathcal{D}_2, \ldots, \mathcal{D}_p\}$ denote a sequence of streaming data blocks where $\mathcal{D}_i \triangleq \{\mathbf{X}_{\mathcal{D}_i},\mathbf{y}_{\mathcal{D}_i}\}$ such that $\{\mathbf{f}_{\mathcal{D}_i}\}_i$ are conditionally independent given $\mathbf{W}$ and $\mathbf{u}$. It can then be shown that (Appendix~\ref{app:a}) $\mathbf{R}_1 = \mathbf{K}^{-1}_\mathcal{UU} + \sum_{i=1}^p \mathbf{E}^{(i)}_1$ and $\mathbf{R}_2 = \sum_{i=1}^p \mathbf{E}^{(i)}_2$ where
\begin{eqnarray}
\mathbf{E}^{(i)}_1 = \frac{1}{\sigma^{2}_n}\mathbf{K}^{-1}_\mathcal{UU}\mathbf{C}^i_\mathcal{UU}\mathbf{K}^{-1}_\mathcal{UU} \ \  \text{;}\ \  
\mathbf{E}^{(i)}_2 = \frac{1}{\sigma^2_n}\mathbf{K}^{-1}_\mathcal{UU}\mathbf{C}_{\mathcal{UD}_i}\mathbf{y}_{\mathcal{D}_i}
\label{eq:3.5}
\end{eqnarray}
where $\mathbf{C}^i_\mathcal{UU} \triangleq \mathbb{E}_{q(\mathbf{W})}[\mathbf{K}_{\mathcal{UD}_i}\mathbf{K}_{\mathcal{D}_i\mathcal{U}}]$ and $\mathbf{C}_{\mathcal{UD}_i} \triangleq \mathbb{E}_{q(\mathbf{W})}[\mathbf{K}_{\mathcal{UD}_i}]$, with $\mathbf{K}_{\mathcal{D}_i\mathcal{U}}$ and  $\mathbf{K}_{\mathcal{U}\mathcal{D}_i}$ being defined similarly to $\mathbf{K}_{\mathcal{D}\mathcal{U}}$ and  $\mathbf{K}_{\mathcal{U}\mathcal{D}}$, respectively (by replacing $\mathcal{D}$ with $\mathcal{D}_i$). Supposing $\mathrm{q}(\mathbf{W})$ is fixed, Eq.~\eqref{eq:3.5} reveals an efficient online update for $\mathrm{q}(\mathbf{u})$ where each update only scales with the size of an incoming data block. Specifically, let $\mathbf{R}^{(i)} = [\mathbf{R}^{(i)}_1; \mathbf{R}^{(i)}_2]$ denote the representation of $\mathrm{q}(\mathbf{u})$ following the arrival of $\{\mathcal{D}_1,\ldots, \mathcal{D}_i\}$ and $\mathbf{E}^{(i+1)} \triangleq [\mathbf{E}^{(i+1)}_1;\mathbf{E}^{(i+1)}_2]$ denote the summary of $\mathcal{D}_{i+1}$,
\begin{eqnarray}
\mathbf{R}^{(i+1)} &=& \mathbf{R}^{(i)} \ +\ \mathbf{E}^{(i+1)} \ .\vspace{-1mm}
\label{eq:3.6}
\end{eqnarray}
This is efficient since the computation of Eq.~\eqref{eq:3.6} only depends on the cost of computing $\mathbf{E}^{(i+1)}$, which in turn only scales linearly with the size of incoming block of data $\mathcal{D}_{i+1}$. If $\mathrm{q}(\mathbf{W})$ is also being updated as data arrives, we would, however, have to recompute $\mathbf{C}_\mathcal{UU}^i$ and $\mathbf{C}_{\mathcal{UD}_i}$ with respect to the updated $\mathrm{q}(\mathbf{W})$. Eq.~\eqref{eq:3.6} therefore incurs a linear recomputation cost in the size of the accumulating dataset and is no longer efficient when data arrives at high frequency. To sidestep this recomputation inefficiency, we instead approximate  $\mathbf{C}^i_\mathcal{UU} \simeq \widehat{\mathbf{C}}^i_\mathcal{UU}$ and $\mathbf{C}_{\mathcal{UD}_i} \simeq \widehat{\mathbf{C}}_{\mathcal{UD}_i}$ using a finite set $\mathbf{P} = \{\mathbf{W}_1,\ldots, \mathbf{W}_k\}$ sampled i.i.d. from the prior $\mathrm{p}(\mathbf{W})$ where
\begin{eqnarray}
\widehat{\mathbf{C}}^i_\mathcal{UU} = \frac{1}{k} \sum_{t=1}^k \frac{\mathrm{q}(\mathbf{W}_t)}{\mathrm{p}(\mathbf{W}_t)}\mathbf{K}^{(t)}_{\mathcal{UD}_i}\mathbf{K}^{(t)}_{\mathcal{D}_i\mathcal{U}}\ \ \text{;}\ \ \widehat{\mathbf{C}}_{\mathcal{UD}_i} = \frac{1}{k} \sum_{t=1}^k \frac{\mathrm{q}(\mathbf{W}_t)}{\mathrm{p}(\mathbf{W}_t)}\mathbf{K}^{(t)}_{\mathcal{UD}_i}
\label{eq:3.7}
\end{eqnarray}
where $\mathbf{K}^{(t)}_{\mathcal{UD}_i}$ and $\mathbf{K}^{(t)}_{\mathcal{D}_i\mathcal{U}}$ denote the covariance matrices evaluated with parameter sample $\mathbf{W}_t$. Since $\mathbf{P}$ can be generated \emph{a priori}, the terms $\{\mathbf{K}^{(t)}_{\mathcal{UD}_i}\mathbf{K}^{(t)}_{\mathcal{D}_i\mathcal{U}},\mathbf{K}^{(t)}_{\mathcal{UD}_i}\}_t$ can be precomputed and cached once $\mathcal{D}_i$ arrives for all future uses. This helps to reduce the recomputation cost of $\mathbf{C}_\mathcal{UU}^i$ and $\mathbf{C}_{\mathcal{UD}_i}$ from $\mathcal{O}(|\mathcal{D}_i|)$ to $\mathcal{O}(k)$ (treating $m$ as a constant). Using Eq.~\eqref{eq:3.7}, we can approximate $\mathbf{E}^{(i)}$, as:
\begin{eqnarray}
{\mathbf{E}}^{(i)}_1 \simeq \widehat{\mathbf{E}}^{(i)}_1 \ =\ \frac{1}{\sigma_n^2}\mathbf{K}^{-1}_\mathcal{UU}\widehat{\mathbf{C}}^i_\mathcal{UU}\mathbf{K}^{-1}_\mathcal{UU} \ \ \text{;}\ \ 
{\mathbf{E}}^{(i)}_2 \simeq \widehat{\mathbf{E}}^{(i)}_2 \ =\  \frac{1}{\sigma_n^2}\mathbf{K}^{-1}_\mathcal{UU}\widehat{\mathbf{C}}_{\mathcal{UD}_i}\mathbf{y}_{\mathcal{D}_i} \ .
\label{eq:3.5b}
\end{eqnarray} 
The streaming update in Eq.~\eqref{eq:3.6} can then be approximated by $\widehat{\mathbf{R}}^{(i+1)} = \widehat{\mathbf{R}}^{(i)} + \widehat{\mathbf{E}}^{(i+1)}$. Supposing all $p$ blocks of data have arrived, this operation incurs only $\mathcal{O}(kp)$ computation cost, which is independent of the number of data points. Furthermore, an appropriate choice of $k$ will guarantee an arbitrarily small approximation loss (Section~\ref{analysis}, Lemma~\ref{lem1}). This is possible via our choices of $\widehat{\mathbf{C}}^i_\mathcal{UU}$ and $\widehat{\mathbf{C}}_{\mathcal{UD}_i}$ in Eq.~\eqref{eq:3.7} which are always unbiased estimates of ${\mathbf{C}}^i_\mathcal{UU}$ and ${\mathbf{C}}_{\mathcal{UD}_i}$.\vspace{-2mm}

\subsection{Online Update for Hyperparameters}
\label{qw}
Following the above update of $\mathrm{q}(\mathbf{u})$, we need to update $\mathrm{q}(\mathbf{W})$ to incorporate the statistical information of the new block of data. Naively, this can be achieved via gradient ascent $\theta \leftarrow \theta + \partial\mathrm{L}(\mathrm{q})/\partial\theta$. This is, however, inefficient as the gradient $\partial\mathrm{L}(\mathrm{q})/\partial\theta$ needs to be re-computed with respect to the entire accumulated dataset as well as the updated $\mathrm{q}(\mathbf{u})$. To sidestep this computational issue, we first notice an additive decomposition (across different blocks of data) of the variational lower-bound. That is, supposing the data stream consists of $N$ data blocks $\{\mathcal{D}_1,\mathcal{D}_2, \ldots,\mathcal{D}_N\}$ of which the agent has received $\mathrm{t}$ data blocks in uniformly random order with $\mathcal{D}_\ast$ being the last block, it follows that (Appendix~\ref{app:b}) $\mathrm{L}({\mathrm{q}}) = \sum_{i=1}^N \mathrm{L}_{\mathcal{D}_i}(\mathrm{q}) - \mathrm{D_{KL}}(\mathrm{q}(\mathbf{u,W})\|\mathrm{p}(\mathbf{u,W}))$
where $\mathrm{L}_{\mathcal{D}_i}(\mathrm{q}) \triangleq \mathbb{E}_{\mathrm{q}(\mathbf{u,W})}[\mathbb{E}_{\mathrm{p}(\mathbf{f}_{\mathcal{D}_i}|\mathbf{u,W})}[\mathrm{log \ p}(\mathbf{y}_{\mathcal{D}_i}|\mathbf{f}_{\mathcal{D}_i})]]$ and $\mathcal{D}_\ast$ can be treated as a random block sampled uniformly from the stream of data $\{\mathcal{D}_1,\mathcal{D}_2, \ldots,\mathcal{D}_N\}$. Using $\mathcal{D}_\ast$, we can construct an unbiased stochastic gradient $\partial\widehat{\mathrm{L}}(\mathrm{q})/\partial\theta$ of $\mathrm{L}(\mathrm{q})$ which satisfies $\mathbb{E}_{\mathcal{D}_\ast}[\partial\widehat{\mathrm{L}}(\mathrm{q})/\partial\theta] = \partial{\mathrm{L}}(\mathrm{q})/\partial\theta$ (Appendix~\ref{app:c}) and is more computationally efficient than the exact gradient $\partial{\mathrm{L}}(\mathrm{q})/\partial\theta$. The computation of $\partial\widehat{\mathrm{L}}(\mathrm{q})/\partial\theta$ only involves $\mathcal{D}_\ast$ and as such, its complexity depends on $|\mathcal{D}_\ast|$ instead of the entire accumulated dataset if we were to use the exact gradient. The resulting stochastic gradient ascent is guaranteed to converge to a local optima given an appropriate schedule of learning rates \cite{Monro1951}. Even though the stochastic gradient above only makes use of the latest block of data $\mathcal{D}_\ast$, the information from previously received data have been extracted and succinctly summarized by the updated $\mathrm{q}(\mathbf{u})$.

{\bf Remark 2.} There also exists other recently developed online GP paradigms such as \cite{Opper02,Bui17} but their representations are not suitable to facilitate communication between agents operating in related domains with different variation scales. In contrast, our developed GP representation characterizes the transformation of the GP prior/posterior from an arbitrary domain to that of a common unit-scale domain and vice versa, thus allowing efficient agent communication across different domains.

\section{Model Fusion}
\vspace{-2mm}
\label{fusion}
This section presents a novel fusion operator which allows two agents to exchange and fuse their local predictive models efficiently (Section~\ref{pairwise}). The resulting operator is generalized to a large-scale model fusion paradigm (Section~\ref{distributed}).\vspace{-2mm}
\subsection{Pairwise Agent Fusion}
\vspace{-2mm}
\label{pairwise}
Suppose two agents learning from two data streams $\mathcal{D}_a \triangleq \{\mathcal{D}_1^a,\ldots \mathcal{D}_{n_a}^{a}\}$ and $\mathcal{D}_b \triangleq \{\mathcal{D}_1^b,\ldots \mathcal{D}_{n_b}^{b}\}$ are respectively characterized by local approximate posteriors $\mathrm{q_a}(\mathbf{u},\mathbf{W}_a) \simeq \mathrm{p}(\mathbf{u},\mathbf{W}_a|\mathbf{y}_{\mathcal{D}_a})$ and $\mathrm{q_b}(\mathbf{u},\mathbf{W}_b) \simeq \mathrm{p}(\mathbf{u},\mathbf{W}_b|\mathbf{y}_{\mathcal{D}_b})$. Since $\mathbf{W}_a$ and $\mathbf{W}_b$ will be marginalized out for prediction, we are interested in approximating the marginal posterior $\mathrm{p}(\mathbf{u}|\mathbf{y}_{\mathcal{D}_a}, \mathbf{y}_{\mathcal{D}_b})$ directly. To achieve this, note that $\mathrm{p}(\mathbf{u}|\mathbf{y}_{\mathcal{D}_a},\mathbf{y}_{\mathcal{D}_b}) \propto \mathrm{p}(\mathbf{u}|\mathbf{y}_{\mathcal{D}_a})\mathrm{p}(\mathbf{u}|\mathbf{y}_{\mathcal{D}_b})/\mathrm{p}(\mathbf{u}) \simeq \mathrm{q}_a(\mathbf{u})\mathrm{q}_b(\mathbf{u}) / \mathrm{p}(\mathbf{u})$
where the first step is shown in Appendix~\ref{app:d}. This implies approximating $\mathrm{p}(\mathbf{u}|\mathbf{y}_{\mathcal{D}_a}, \mathbf{y}_{\mathcal{D}_b})$ can be achieved via constructing the fusion statistics $\mathrm{q}_{ab}(\mathbf{u}) \propto {\mathrm{q}_a(\mathbf{u})\mathrm{q}_b(\mathbf{u})}/{\mathrm{p}(\mathbf{u})}$. Specifically, let $\mathrm{q}_a(\mathbf{u}) = \mathcal{N}(\mathbf{u}|\mathbf{m}_a, \mathbf{S}_a)$ and $\mathrm{q}_b(\mathbf{u}) = \mathcal{N}(\mathbf{u}|\mathbf{m}_b, \mathbf{S}_b)$ where the parameters $\mathbf{m}_a,\mathbf{m}_b,\mathbf{S}_a$, and $\mathbf{S}_b$ are computed using Eq.~\eqref{eq:2.4}. Then $\mathrm{q}_{ab}(\mathbf{u}) = \mathcal{N}(\mathbf{u}|\mathbf{m}_{ab}, \mathbf{S}_{ab})$ where (Appendix~\ref{app:e}):
\begin{eqnarray}
\mathbf{S}_{ab} \ =\ \left(\mathbf{S}^{-1}_a + \mathbf{S}^{-1}_b - \mathbf{K}_{\mathcal{UU}}^{-1}\right)^{-1} \ \text{;}\ \  
\mathbf{m}_{ab} \ =\ \mathbf{S}_{ab} \left(\mathbf{S}^{-1}_a \mathbf{m}_a + \mathbf{S}^{-1}_b \mathbf{m}_b \right) \ .
\label{eq:4.10}
\end{eqnarray}
Let $\mathbf{R}_{ab}$, $\mathbf{R}_a$, $\mathbf{R}_b$, and $\mathbf{R}_0$ respectively be the natural representation of
$\mathrm{q}_{ab}(\mathbf{u})$, $\mathrm{q}_a(\mathbf{u})$, $\mathrm{q}_b(\mathbf{u})$, and $\mathrm{p}(\mathbf{u})$ (see Section~\ref{qu}). Eq.~\eqref{eq:4.10} can be rewritten concisely as $\mathbf{R}_{ab} = \mathbf{R}_{a} + \mathbf{R}_{b} - \mathbf{R}_{0}$. In practice, however, since maintaining $\mathbf{R}_a$ and $\mathbf{R}_b$ is not efficient for online update, we instead use their approximated versions $\widehat{\mathbf{R}}_a$ and $\widehat{\mathbf{R}}_b$ (see Section~\ref{qu}) to approximate $\mathbf{R}_{ab}$ by $\widehat{\mathbf{R}}_{ab} = \widehat{\mathbf{R}}_{a} + \widehat{\mathbf{R}}_{b} - \mathbf{R}_{0}$. This fusion operator's total cost depends only on the size of $\mathbf{u}$ and is constant w.r.t data size.

{\bf Remark 3.} Although $\mathrm{q}(\mathbf{W}_a)$ and $\mathrm{q}(\mathbf{W}_b)$ are not fused explicitly, they will still be updated later using $\mathrm{q}(\mathbf{u})$ when new data arrives (see Remark $2$). This implicitly helps agents utilizing the fused model to improve their projection matrices $\mathbf{W}_a$ and $\mathbf{W}_b$ for better cross-domain mapping (Section~\ref{fgp}). 
\vspace{-5mm}
\subsection{Decentralized Multi-Agent Fusion}
\vspace{-2mm}
\label{distributed}
This section extends the above pairwise fusion protocol to facilitate model fusion beyond two agents. Specifically, consider a distributed network of $s$ independent agents with local models $\mathrm{q}_i(\mathbf{u}) \simeq \mathrm{p}(\mathbf{u} | \mathbf{y}_{\mathcal{D}_i})$ for $1 \leq i \leq s$. Let $\mathbf{R}_1,\mathbf{R}_2,\ldots,\mathbf{R}_s$ denote their exact representations, it can be shown that (Appendix~\ref{app:f}) the representation $\mathbf{R}_g$ of their fused model $\mathrm{q}(\mathbf{u}) \simeq \mathrm{p}(\mathbf{u} | \mathbf{y}_{\mathcal{D}_1}, \ldots, \mathbf{y}_{\mathcal{D}_s})$ is $\mathbf{R}_g = \sum_{i = 1}^{s} \mathbf{R}_i - (s - 1)\mathbf{R}_0$ where $\mathbf{R}_0$ denotes the natural representation of prior $\mathrm{p}(\mathbf{u})$.

Naively, $\widehat{\mathbf{R}}_g$ can be approximated by $\widehat{\mathbf{R}}_g = \sum_{i = 1}^{s} \widehat{\mathbf{R}}_i - (s - 1)\mathbf{R}_0$ using $\widehat{\mathbf{R}}_1,\ldots,\widehat{\mathbf{R}}_s$ for efficient online update (Section~\ref{qu}). This, however, requires either direct communication between every two agents or a central server through which agents coordinate their communications. The former implies a fully connected network which is not desirable in situations that require large spatial coverage such as environmental sensing \cite{NghiaICML14} or terrain exploration \cite{LowAAMAS12,LowAAMAS13} while the latter will create a computational bottleneck and risk exposing a single choke point for failure. To avoid these issues, this section develops a decentralized model fusion algorithm that allows agents to exchange local representations as messages among one another within their broadcasting ranges. 


In particular, let $\mathbf{M}^{t +1}_{ij}$ denote the message that agent $i$ sends to agent $j$ (within broadcasting range) at time step $t + 1$, which summarizes and integrates $i$'s local representation with the shared representations it received from other agents in the previous $t$ steps of communication. This must not include the representation of agent $j$ to avoid aggregating duplicates of knowledge. Thus, $\mathbf{M}^{t+1}_{ij}$ should essentially aggregate the representation of all agents (excluding $j$) whose messages can reach $i$ within $t$ steps of direct transmission. 

As such, $\mathbf{M}^{t+1}_{ij}$ can be recursively computed by aggregating only received messages from those in $i$'s local neighborhood in the previous time step $t$, $\mathbf{M}^{t + 1}_{ij} = \widehat{\mathbf{R}}_i + \sum_k (\mathbf{M}^t_{ki} - \mathbf{R}_0)$ where $k \in \mathbb{N}(i)\setminus\{j\}$ and $\mathbb{N}(i)$ denotes the neighborhood of $i$ and the subtraction of $\mathbf{R}_0$ from $\mathbf{M}^t_{ki}$ is to prevent aggregating multiple copies of the prior model's representation $\mathbf{R}_0$, which has already been aggregated into $\widehat{\mathbf{R}}_i$, by definition. At time $t = 0$, the message only contains $i$'s local representation (i.e., $\mathbf{M}_{ij}^t = \widehat{\mathbf{R}}_i$) since obviously, only $i$ can reach itself in $0$ step of transmission. Upon convergence at $t = t_{\max}$\footnote{For a tree-topology network, the above message passing algorithm will converge to the exact optimum after $t_{\max}$ time-steps where $t_{\max}$ is the tree's diameter. The agents can employ decentralized minimum spanning tree to eliminate redundant connections with high latencies to guarantee that their connection topology is a tree.}, each agent $i$ can aggregate the received messages to assemble the same global representation, $\widehat{\mathbf{R}}_g = \widehat{\mathbf{R}}_i + \sum_k (\mathbf{M}^{t_{\max}}_{ki} - \mathbf{R}_0)$ where $k \in \mathbb{N}(i)$ and again, the repeated subtraction of $\mathbf{R}_0$ from $\mathbf{M}^{t_{\max}}_{ki}$ is to prevent aggregating multiple copies of $\mathbf{R}_0$ into $\widehat{\mathbf{R}}_g$. 

\vspace{-3mm}
\section{Theoretical Analysis}
\vspace{-2mm}
\label{analysis}
This section shows that the approximate global approximation can be made arbitrarily close to the exact representation $\mathbf{R}_g$ with high confidence (Theorem~\ref{theo1}). In particular, we are interested in bounding the difference between $\mathbf{R}_g$ and its approximation $\widehat{\mathbf{R}}_g$ w.r.t the numbers $k$ of projection matrices, $s$ of agents and the size $m$ of the encoding vocabulary. Let $\mathbf{R}_i$ be the exact representation for agent $i$ and $\widehat{\mathbf{R}}_i$ be its approximation generated by our framework (Section~\ref{qu}), the difference between $\mathbf{R}_i$ and $\widehat{\mathbf{R}}_i$ is bounded below:

\begin{lemma}[Representation Loss]
	Given $\epsilon > 0$ and $\delta \in (0,1)$, it can be guaranteed that with probability at least $1 - \delta$, $\|\mathbf{R}_i - \widehat{\mathbf{R}}_i\| \leq \epsilon$ by choosing $k = \displaystyle\mathcal{O}((m^2/\epsilon^2)\mathrm{log}(m/\delta))$.
	\label{lem1}
\end{lemma}
{\bf Proof.} A detailed proof is provided in Appendix~\ref{app:g}.

Exploiting the result of Lemma~\ref{lem1}, we can bound the difference between $\mathbf{R}_g$ and $\widehat{\mathbf{R}}_g$ with high probability in terms of $m$, $s$, and $k$, as detailed in Theorem~\ref{theo1} below.

\begin{theorem}[Fusion Loss]
	\label{theo1}
	Given $\epsilon > 0$ and $\delta \in (0,1)$, it can be guaranteed that with probability at least $1 - \delta$, $\|\mathbf{R}_g - \widehat{\mathbf{R}}_g\| \leq \epsilon$ by choosing $k = \displaystyle\mathcal{O}((m^2s^2/\epsilon^2)\mathrm{log}(ms/\delta))$.
\end{theorem}
{\bf Proof.} A detailed proof is provided in Appendix~\ref{app:h}.

{\bf Remark $3$.} The above results imply that both the representation and fusion losses can be made arbitrarily small with high probability by choosing a sufficiently large number of cross-domain projection matrix samples (Section~\ref{qu}) to approximately represent each agent's predictive model. In addition, Theorem~\ref{theo1} also tells us that the no. of samples $k$ needs to grow quadratically in the size of the encoding vocabulary and the no. of agents to guarantee the above. This means the agent's complexity needs to increase to guarantee fusion quality when we have more agents.\vspace{-1mm}
\vspace{-3mm}
\section{Experiments}
\label{exp}
This section demonstrates our decentralized  \underline{Co}llective \underline{O}nline \underline{L}earning {GP} (COOL-GP) framework's efficiency, resiliency to information disparity, and fault-tolerance to information loss on several synthetic and real-world domains:

(a) The SYNTHETIC domain features two streaming datasets generated by $\mathrm{f}_1(\mathbf{x}) \triangleq \mathrm{u}(\mathbf{W}_1\mathbf{x})$ and $\mathrm{f}_2(\mathbf{x}) \triangleq \mathrm{u}(\mathbf{W}_2\mathbf{x})$ where the common random function $\mathrm{u}(\mathbf{z})$ is sampled from a standardized GP (Section~\ref{fgp}) with different projection matrices $\mathbf{W}_1$ and $\mathbf{W}_2$. Each dataset comprises of $200$ batches of $6$-dimensional training data which amount to $8000$ data points. A separate dataset of $4000$ data points (generated from both $\mathrm{f}_1$ and $\mathrm{f}_2$) is used for testing. 

(b) The AIRLINE domain \cite{Hensman13,NghiaICML15} features an air transportation delay phenomenon that generates a stream of data comprising of $30000$ batches of observations ($600000$ data points in total). Each batch consists of $20$ observations. Each observation is a $8$-dimensional feature vector containing the information log of a commercial flight and a corresponding output recording its delay time (min). The system comprises of $1000$ agents. Each agent is tested on a separate set of $10000$ data points. 

(c) The AIMPEAK domain \cite{NghiaICML16} features a traffic phenomenon which took place over an urban road network comprising of $775$ road segments. $10000$ batches of data are then generated from the traffic phenomenon and streamed in random order to a group of $100$ collective learning agents. Each observation is a $5$-dimensional input vector. Its output corresponds to the traffic speed (km/h). The predictive performance of each agent is then evaluated using a separate test set of $2000$ data points.

In all experiments, each data batch arrives sequentially in a random order and is dispatched to a random learning agent. This simulates learning scenarios with streaming data where agents collect one batch of data at a time. We report the averaged predictive performance before and after fusion of the agents vs. the number of arrived batches of data to demonstrate the efficiency of our collective learning paradigm in such distributed data streaming settings as a proof-of-concept.



\begin{figure}[t]
	\begin{tabular}{ccc}
		\hspace{-2mm}\includegraphics[width=4.2cm]{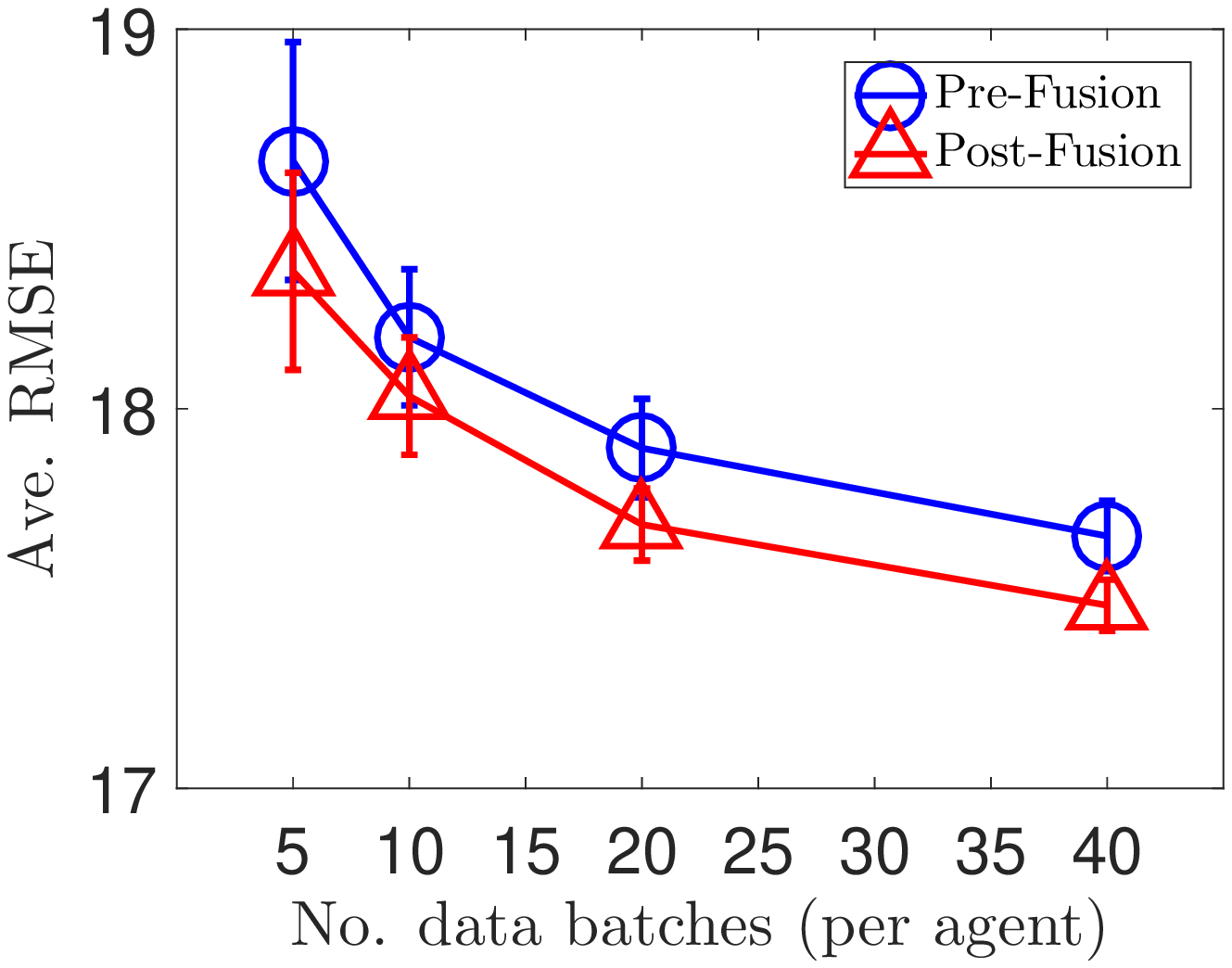} & \includegraphics[width=4.2cm]{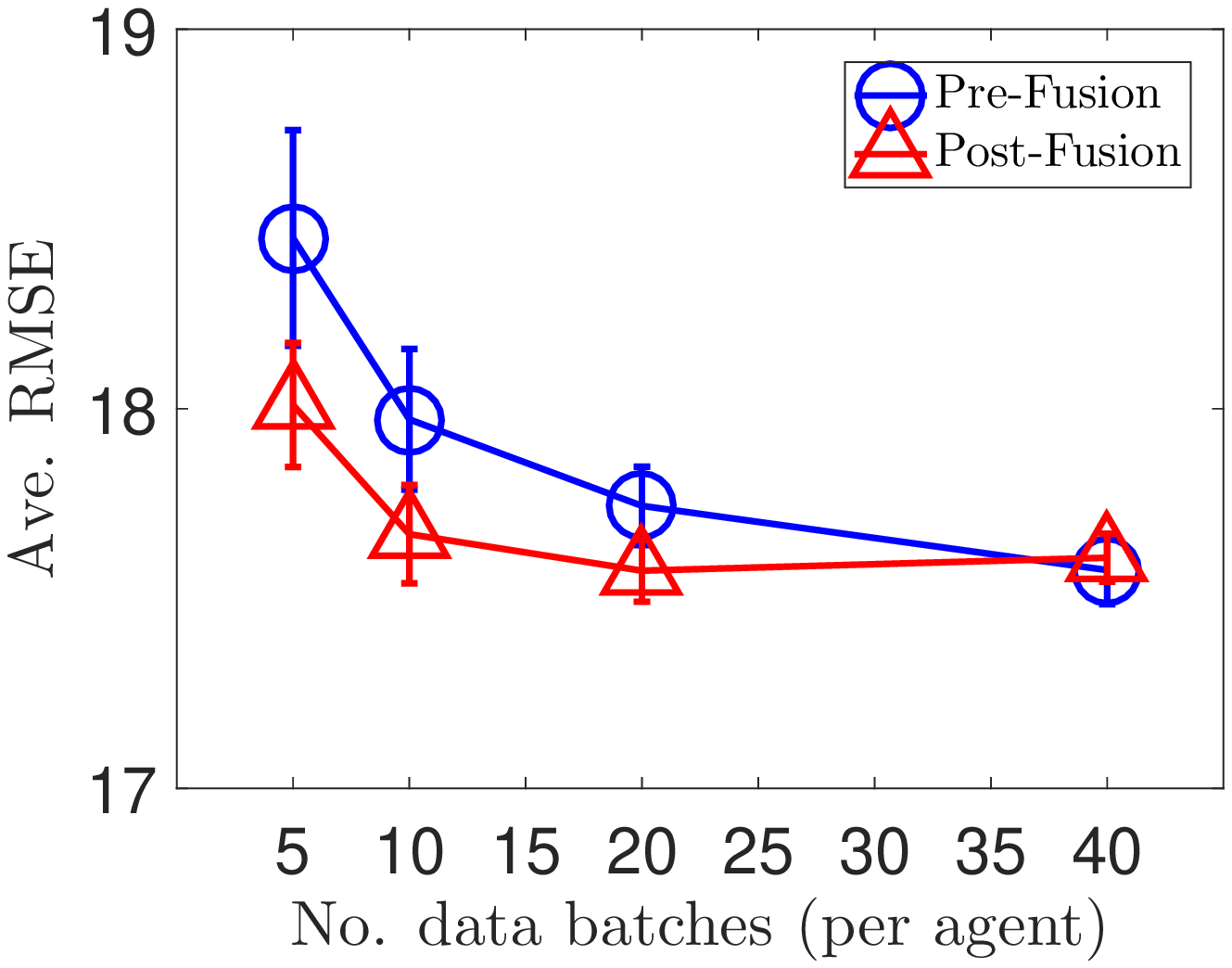} & \includegraphics[width=4.2cm]{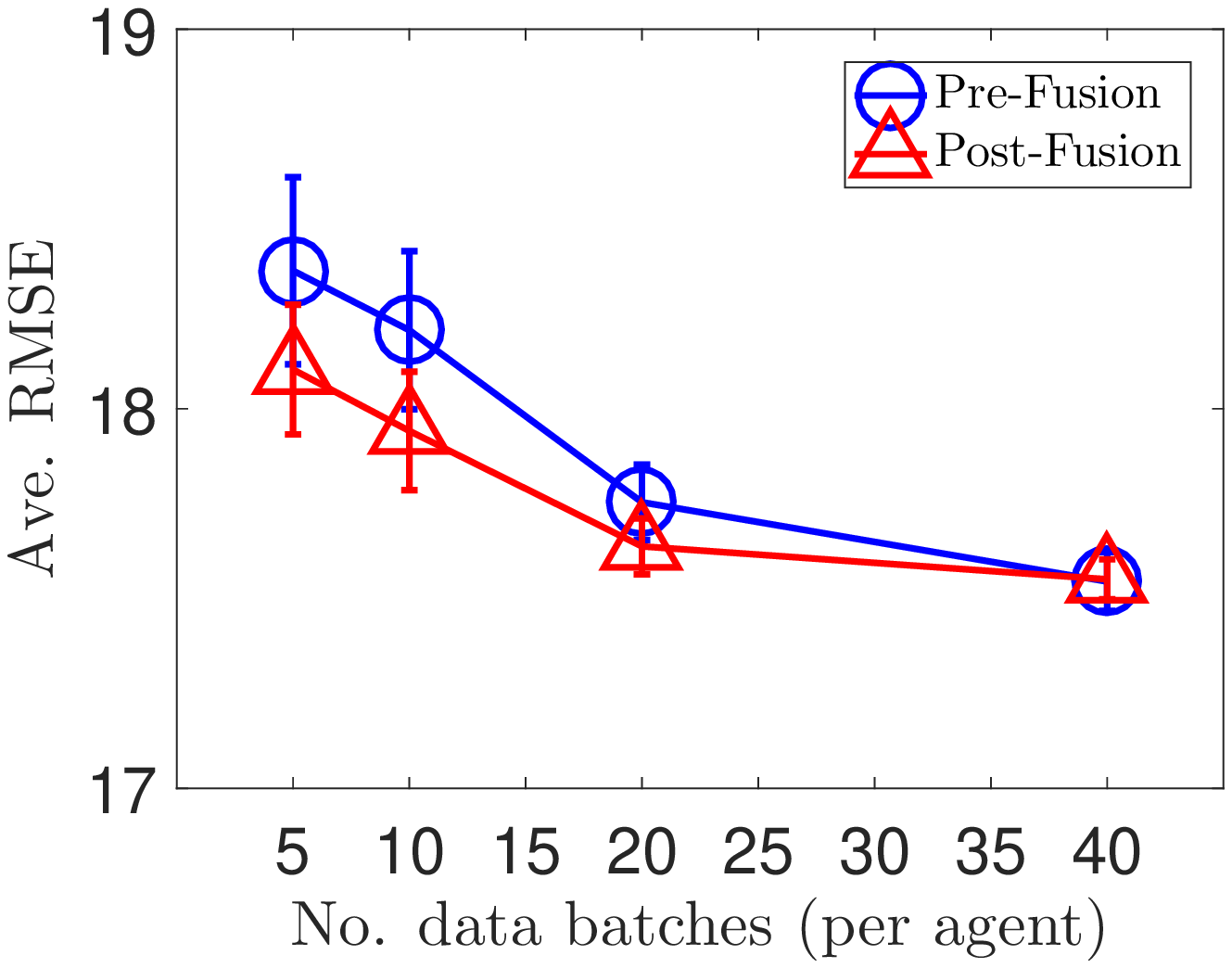}\\
		\hspace{1mm}(a) $|\mathbf{Z}| = 50$ $\&$ $|\mathbf{P}| = 5$ & (b) $|\mathbf{Z}| = 100$ $\&$ $|\mathbf{P}| = 5$ & (c) $|\mathbf{Z}| = 150$ $\&$ $|\mathbf{P}| = 5$\vspace{2mm}\\ 
		\hspace{-2mm}\includegraphics[width=4.2cm]{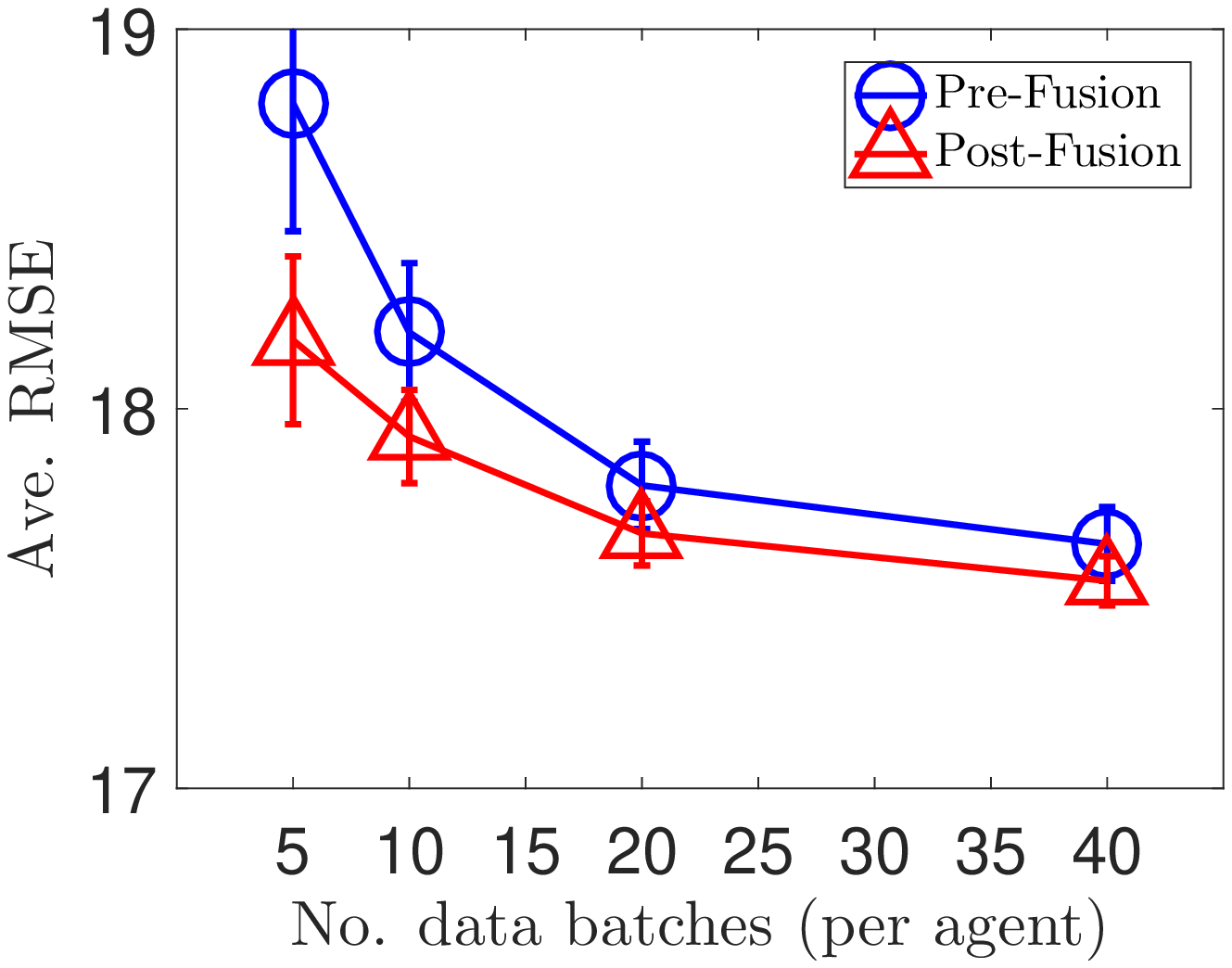} & \includegraphics[width=4.2cm]{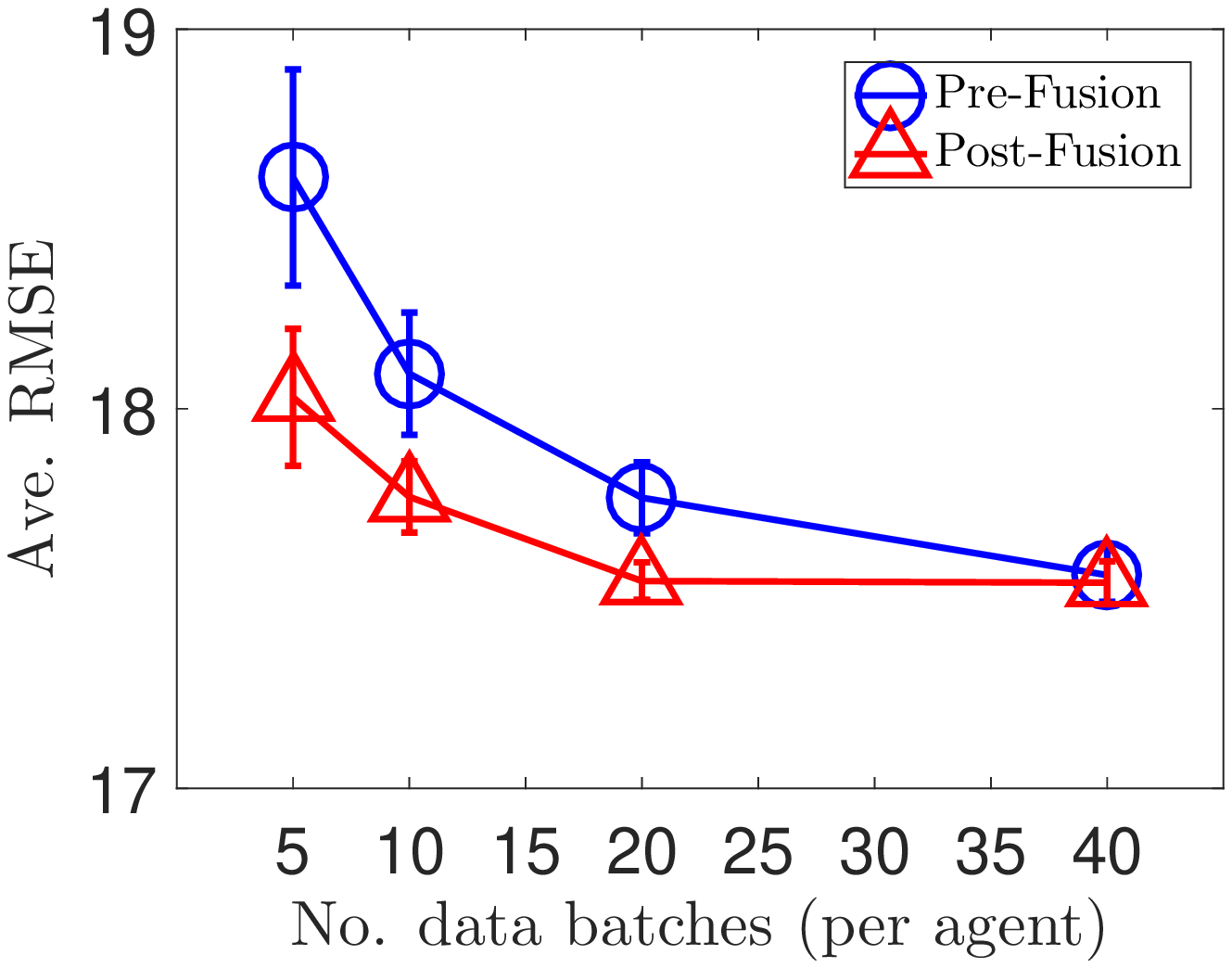} & \includegraphics[width=4.2cm]{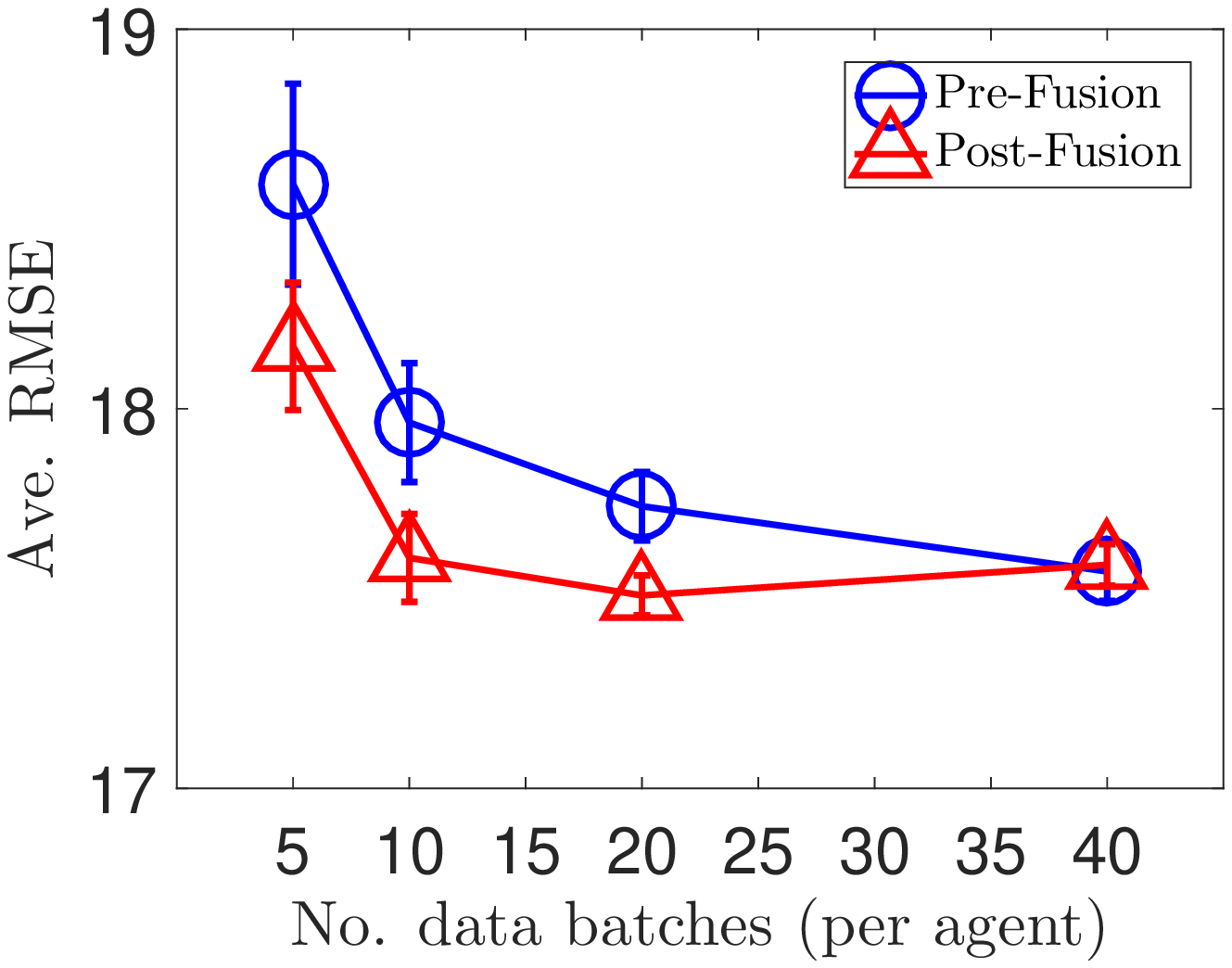}\\
		\hspace{1mm}(d) $|\mathbf{Z}| = 50$ $\&$ $|\mathbf{P}| = 10$ & (e) $|\mathbf{Z}| = 100$ $\&$ $|\mathbf{P}| = 10$ & (f) $|\mathbf{Z}| = 150$ $\&$ $|\mathbf{P}| = 10$
	\end{tabular}
	\caption{Graphs of averaged pre- and post-fusion performance vs. no. of data batches dispatched to $2$ agents with varying sizes of encoding vocabulary $|\mathbf{Z}|$ and projection matrix samples $|\mathbf{P}|$.}
	\label{fig:synthetic}
\end{figure}

\begin{figure}[t]
	\begin{tabular}{ccc}
		\hspace{-2mm}\includegraphics[width=4.2cm]{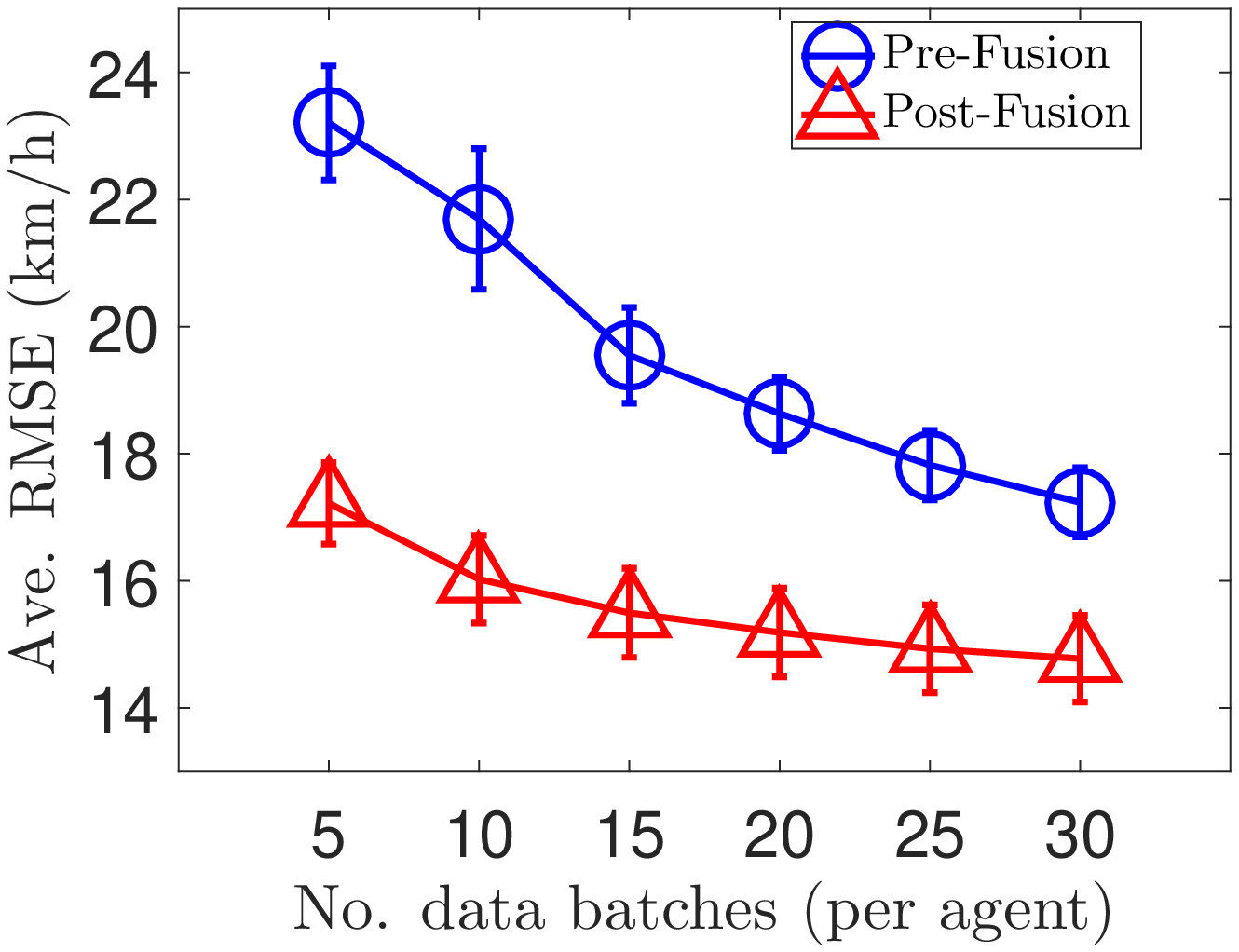} &\includegraphics[width=4.2cm]{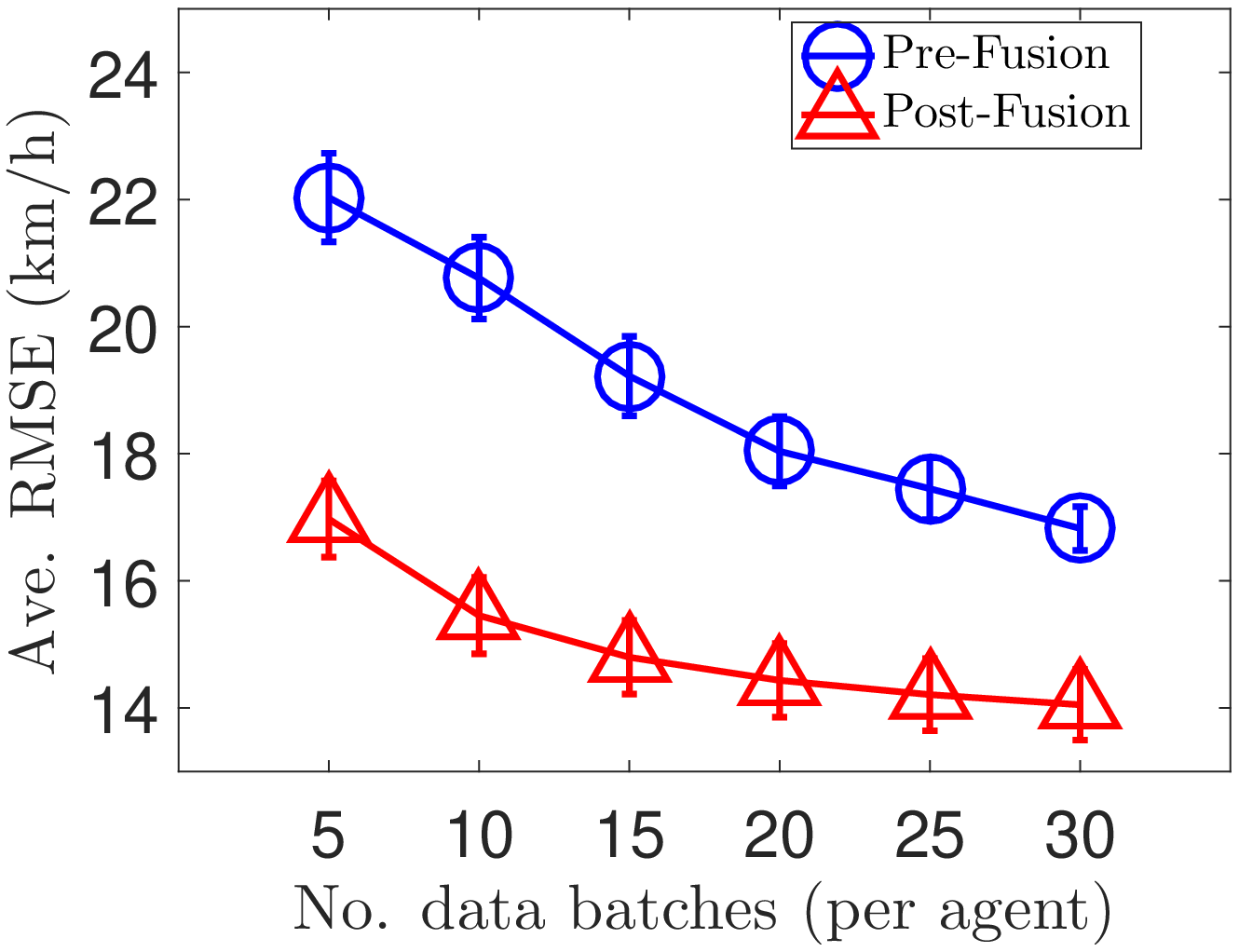} 
		&\includegraphics[width=4.2cm]{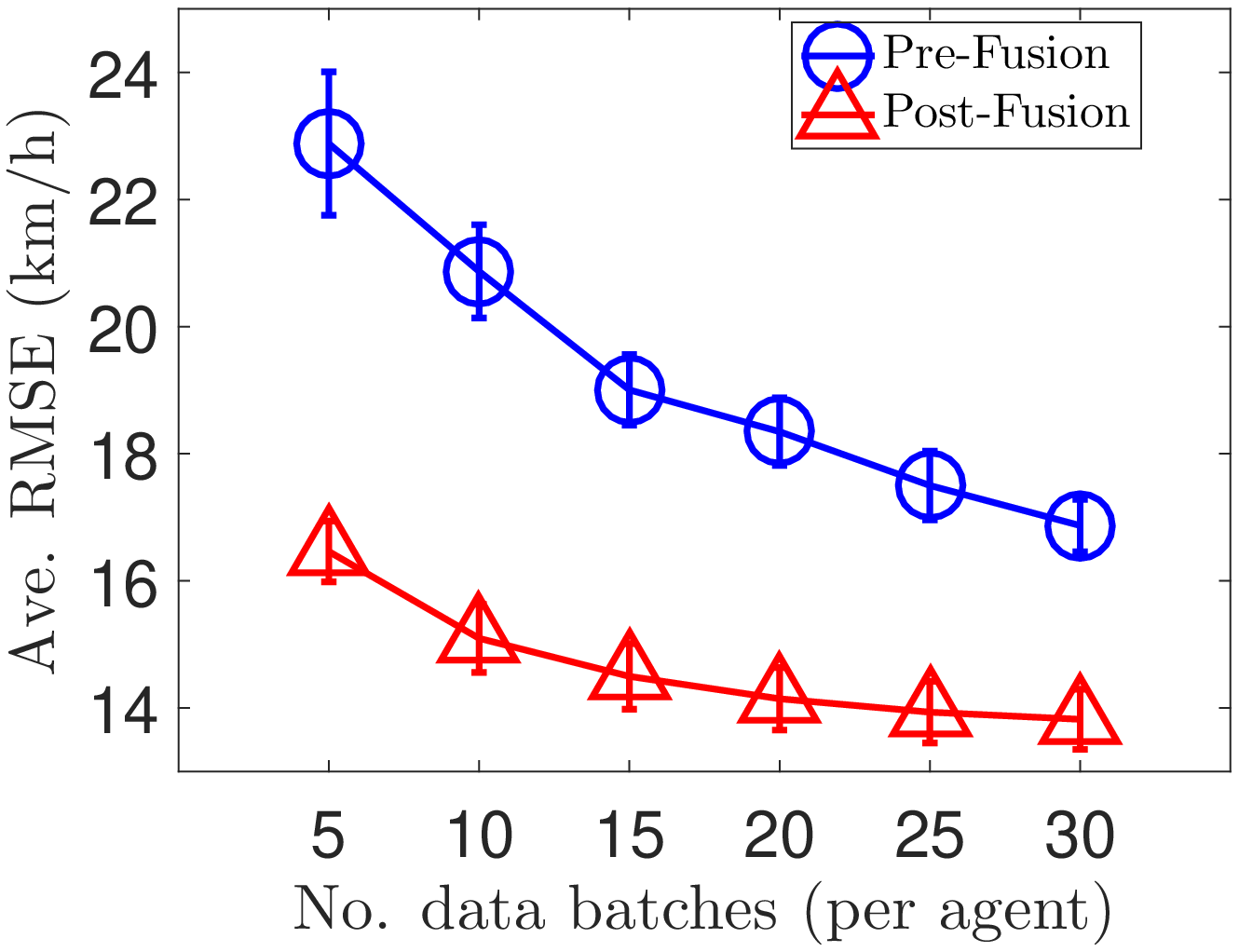}\\
		\hspace{1mm}(a) $|\mathbf{Z}| = 50$ $\&$ $|\mathbf{P}| = 5$ & (b) $|\mathbf{Z}| = 100$ $\&$ $|\mathbf{P}| = 5$ & (c) $|\mathbf{Z}| = 200$ $\&$ $|\mathbf{P}| = 5$\vspace{2mm}\\ 
		\hspace{-2mm}\includegraphics[width=4.2cm]{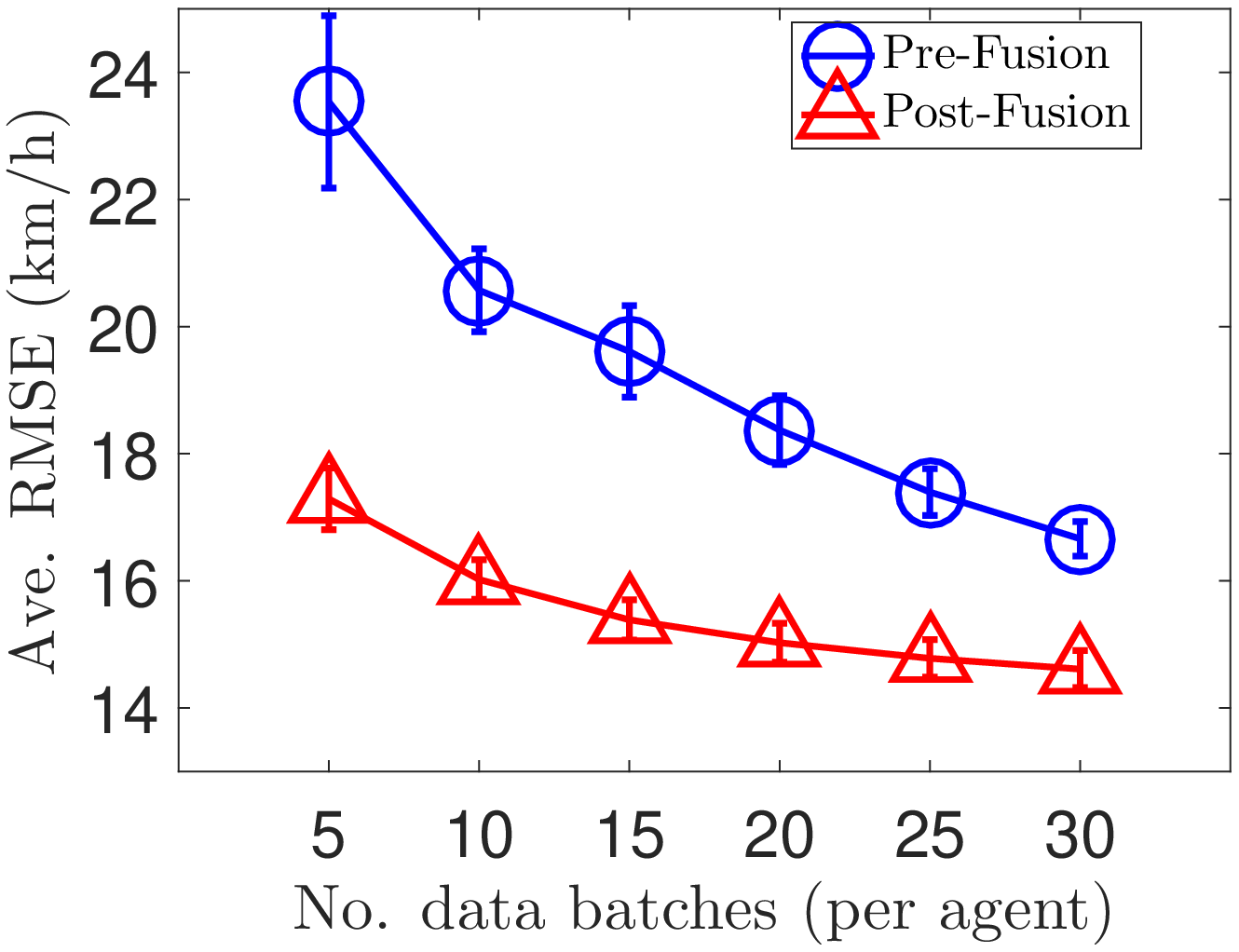} & \includegraphics[width=4.2cm]{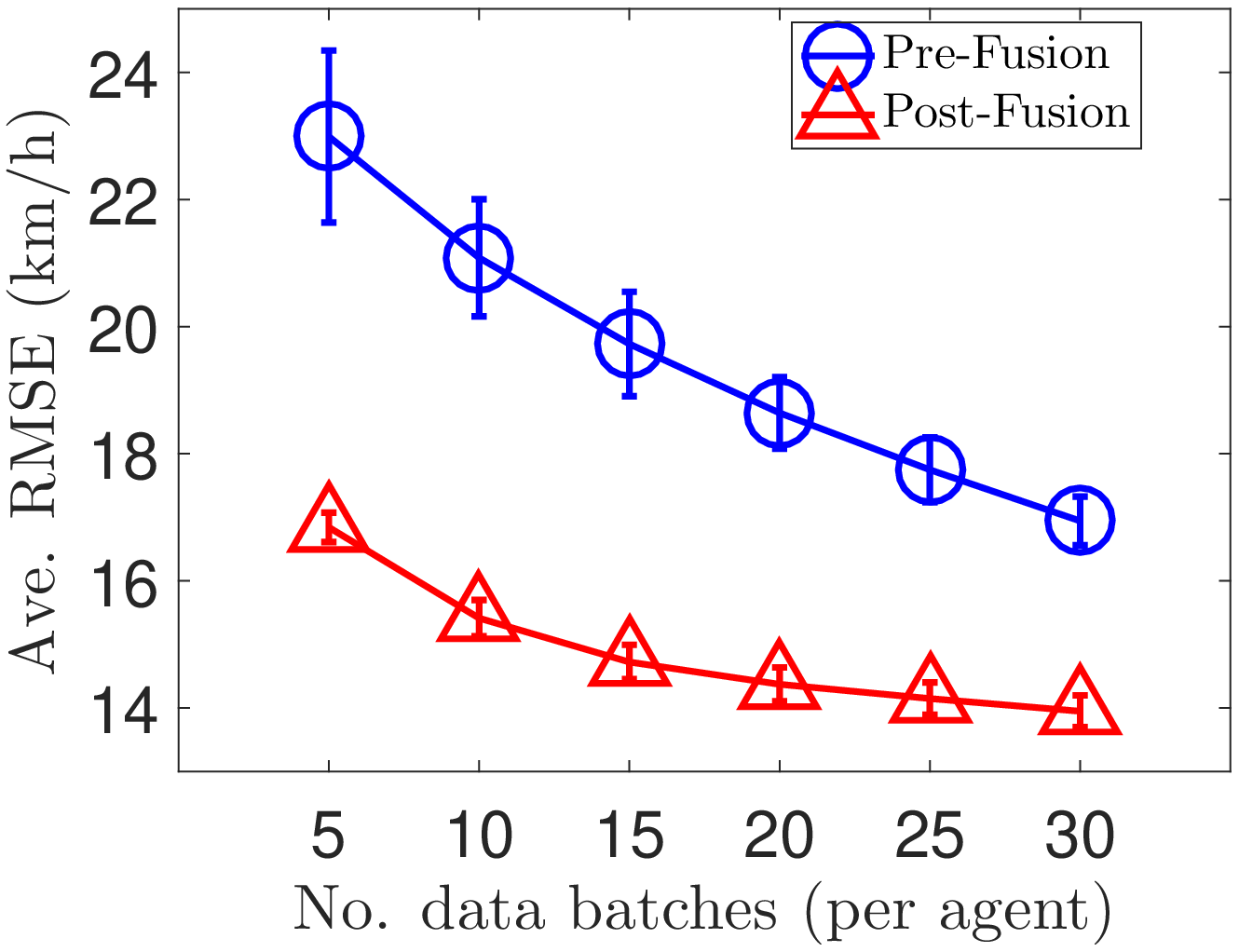} &
		\includegraphics[width=4.2cm]{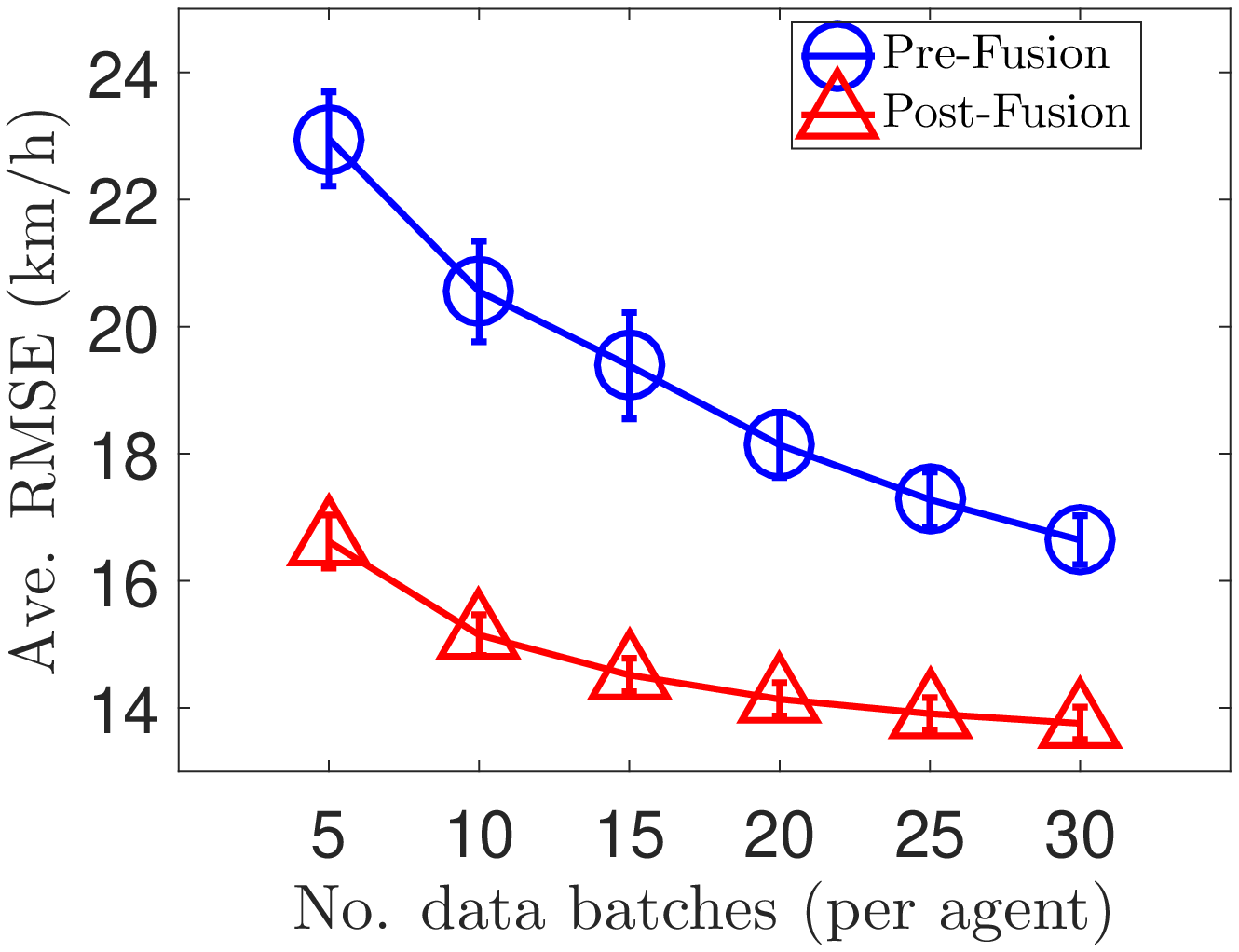} \\
		\hspace{1mm}(d) $|\mathbf{Z}| = 50$ $\&$ $|\mathbf{P}| = 20$ & (e) $|\mathbf{Z}| = 100$ $\&$ $|\mathbf{P}| = 20$ & (f) $|\mathbf{Z}| = 200$ $\&$ $|\mathbf{P}| = 20$ 
	\end{tabular}
	\caption{Graphs of averaged pre- and post-fusion performance vs. no. of data batches of $100$ agents collecting data from the same traffic phenomenon with varying $|\mathbf{Z}|$ and $|\mathbf{P}|$.}
	\label{fig:aimpeak}
\end{figure}
 
Fig.~\ref{fig:synthetic} reports the results of our COOL-GP framework in a cross-domain learning scenario where two agents integrate their predictive models of two correlated, synthetic phenomena to improve their averaged performance on test instances from both domains. Fig.~\ref{fig:aimpeak} further reports the performance of COOL-GP in a real-world traffic monitoring application deployed on a large, decentralized network consisting of $100$ learning agents. Both of these cases demonstrate the effect of COOL-GP fusion on the averaged predictive accuracy w.r.t varying amount of dispatched data batches for different choices of encoding vocabulary sizes $|\mathbf{Z}|$ and the sampling size $|\mathbf{P}|$ used to approximate the agent's representation (Section~\ref{qu}). Across all configurations, a consistent pattern can be observed: (a) post-fusion predictions exhibit significant performance gain as compared to pre-fusion predictions; and (b) the performance gap gradually closes up with more data collected, which suggests a diminishing marginal gain of model fusion. 

\begin{figure}[t]
	\begin{tabular}{ccc}
		\hspace{-3mm}\includegraphics[width=4.5cm]{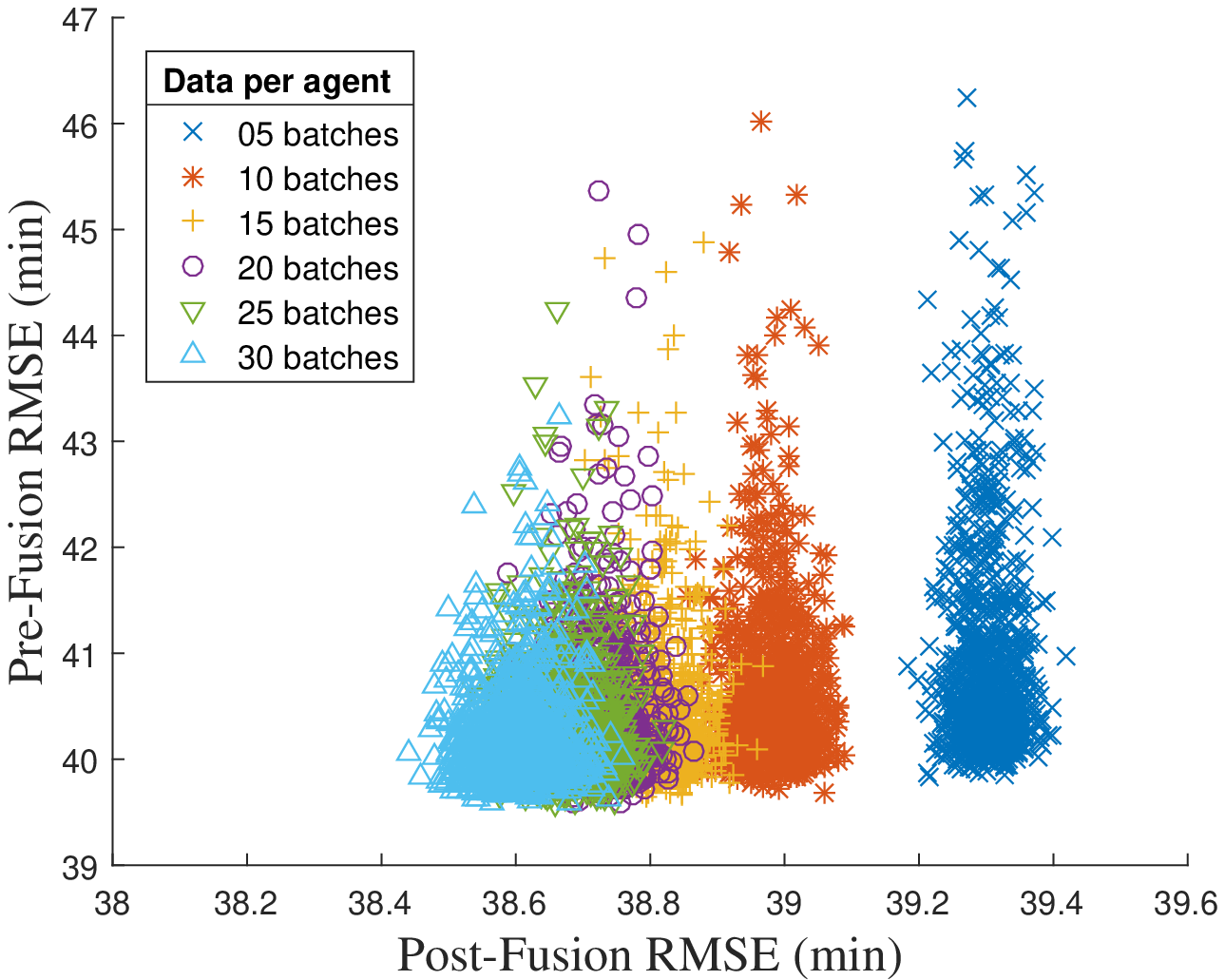} &\hspace{-3mm}
		\includegraphics[width=4.5cm]{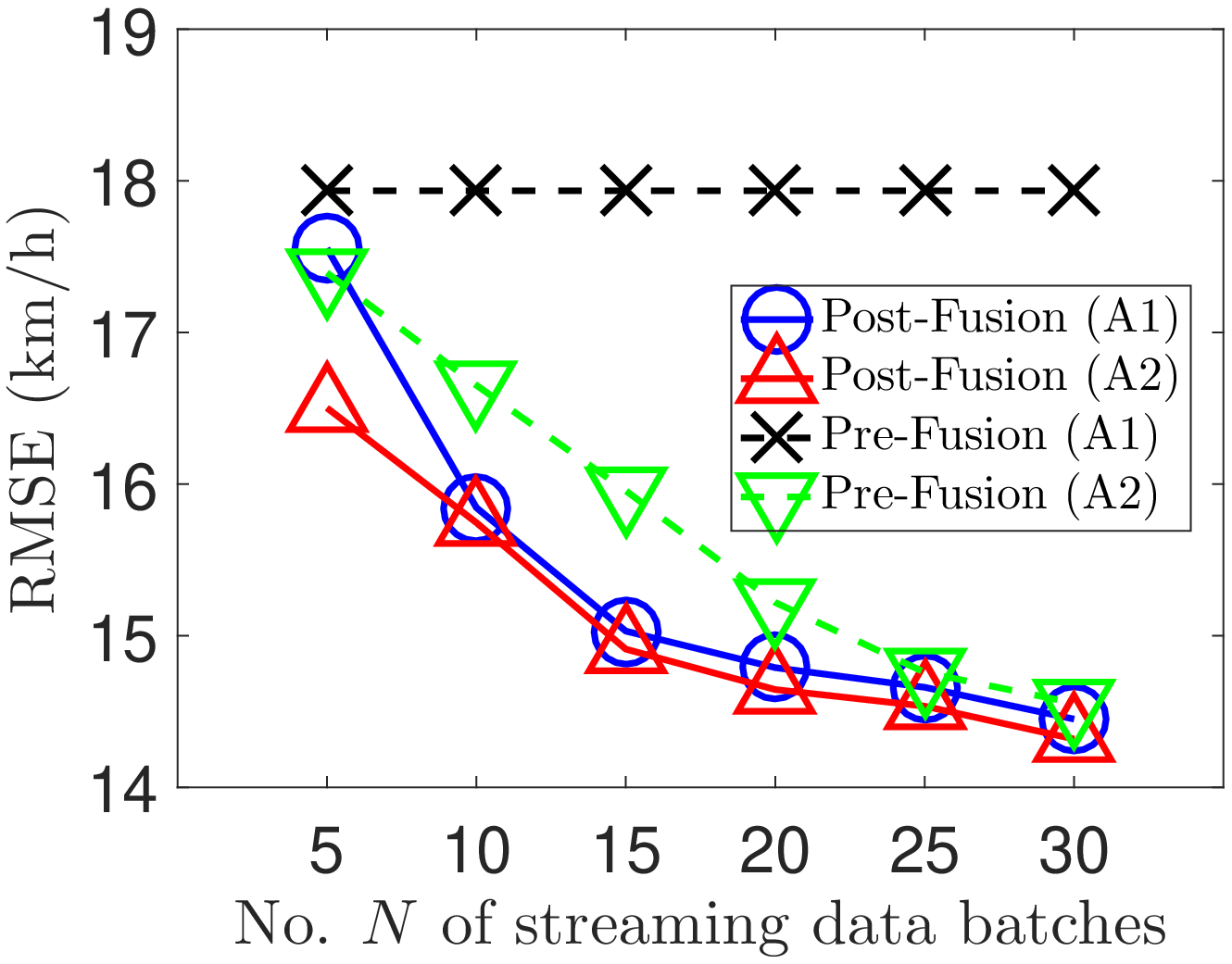} & \hspace{-3mm}
		\includegraphics[width=4.5cm]{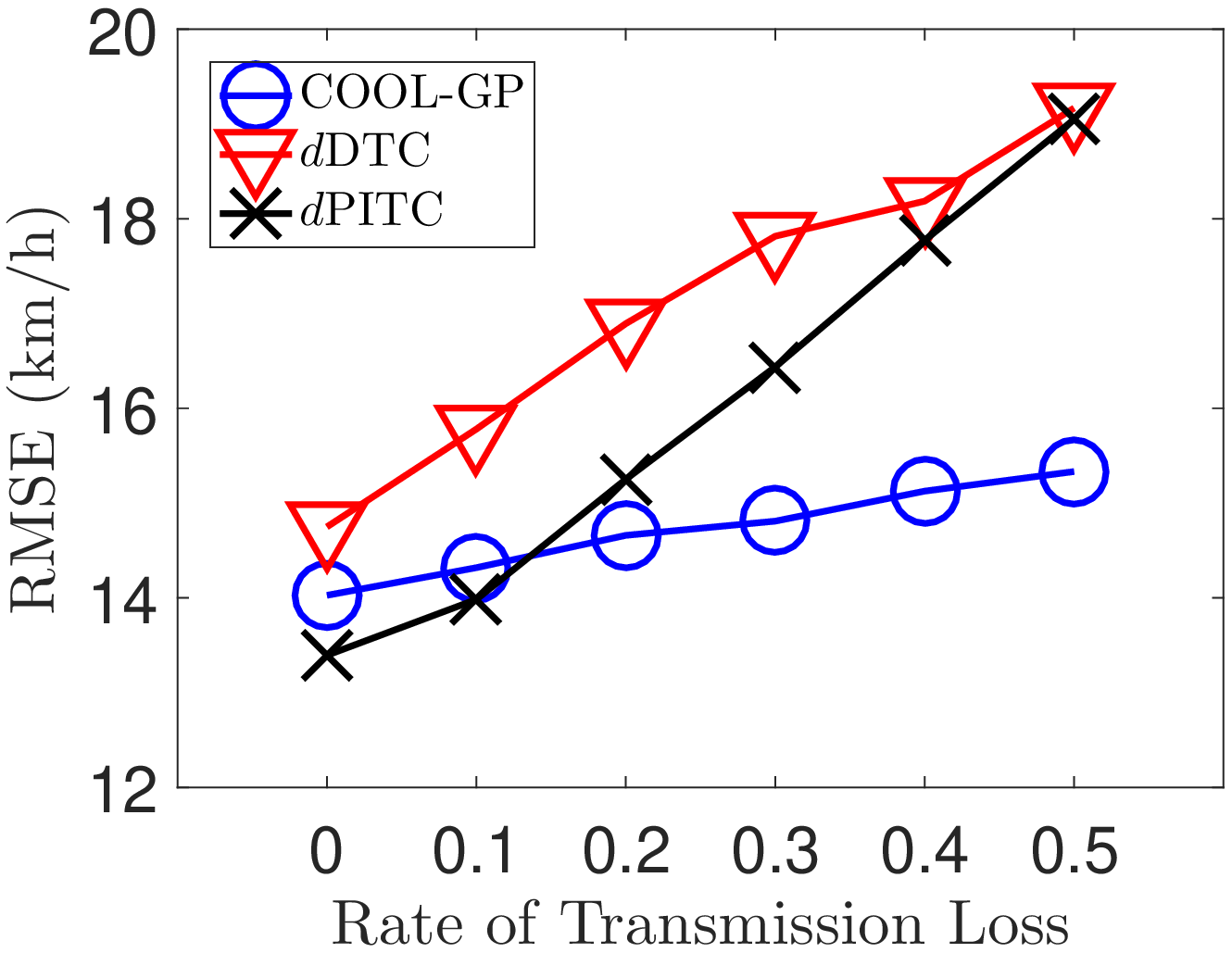} \\
		(a) & (b) & (c) \vspace{-2mm}
	\end{tabular}
	\caption{Graphs of (a) individual performance profiles (pre- vs. post-fusion RMSE) of a $1000$-agent system collectively learning using our COOL-GP framework in the AIRLINE domain \cite{Hensman13,NghiaICML15}; (b) pre- and post-fusion individual performance of two agents with different learning capabilities; and (c) post-fusion performance of COOL-GP in comparison to those of state-of-the-art distributed GPs (e.g., $d$DTC \cite{Yarin14} and $d$PITC \cite{NghiaICML16}) vs. rate of transmission loss in the AIMPEAK domain.}\vspace{-4mm}
	\label{fig:profile}
\end{figure}

Fig.~\ref{fig:profile} visualizes a comprehensive collection of individual performance profiles of $1000$ agents in the AIRLINE domain (each profile is represented by a pair of pre- and post-fusion RMSEs). The result shows that with more data collected, clusters of performance profiles (i.e., each cluster is visualized by a colored point cloud) gradually migrate towards regions with superior pre- and post-fusion accuracy. The migration distance, however, reduces rapidly in latter stages of data collection, which is consistent with the previous observation on the diminishing return of model fusion. Interestingly, it can also be observed that within each cluster, the performance profiles exhibit high variance for pre-fusion and low variance for post-fusion performance, which suggests that agents are able to achieve post-fusion consensus within small range of variation (i.e., fusion stability). \vspace{-1mm}

We also investigate an interesting case study of model fusion between agents allocated with different amounts of data in the AIMPEAK traffic domain. Specifically, Fig.~\ref{fig:profile}b reports the performance of two agents A$1$ (fixed amount of data) and A$2$ (continuous supply of data). Without fusion, A$1$ fails to update its model, and improve its performance as expected, whereas A$2$ still exhibits gain in performance as it receives more data. With fusion, however, the performance of A$1$ is brought close to that of A$2$ and far exceeds its original accuracy. More interestingly, it can be observed that the performance of A$2$ also marginally improves upon fusion with a conservative A$1$ that never collects new data to update its model. This demonstrates that COOL-GP greatly benefits agents with lesser learning capabilities and, at the same time, mildly improves the performance of those with better capabilities (i.e., resiliency to information disparity). 

Finally, in the traffic domain (i.e., AIMPEAK), we present another interesting case study that features a distributed learning scenario among 100 agents where each transmission of local representations (or local statistics in the cases of cloud-oriented distributed GPs such as $d$DTC \cite{Yarin14} and $d$PITC \cite{NghiaICML16}) might not reach its destination with a certain probability. The averaged post-fusion performance are plotted against the rate of transmission loss to demonstrate the high fault-tolerance of our COOL-GP. Fig.~\ref{fig:profile}c shows that, as transmission losses occur more frequently, the averaged performance of COOL-GP agents degrades more gracefully than those of state-of-the-art\footnote{We do not compare with $d$PIC \cite{NghiaICML16} as it requires storing local data and is not suitable for online learning.} distributed learning frameworks $d$DTC and $d$PITC which communicate directly to a central server that coordinates them. This is expected since both $d$DTC and $d$PITC require every agent to successfully transmit its local model directly to a single master server. Failing to achieve this immediately leads to irrecoverable information loss. In contrast, COOL-GP allows each local agent to propagate its model to multiple agents within its neighborhood (see Section~\ref{distributed}), thus lowering the risk of losing information.



\vspace{-4mm}
\section{Conclusion}
\vspace{-2mm}
\label{conclude}
Traditional distributed algorithms for ML implemented with server-client architecture are often undesirable due to the centralized risk of operational failure and various capacity bottlenecks imposed by the server. In this paper, we advocate a shift in paradigm towards distributed ML paradigm with peer-to-peer decentralized communication architecture, which exploits the collective computation capacities of local devices and preserves analytic quality through on-demand integration of local models. Specifically, we propose a collective decentralized Gaussian process (GP) framework that is to be simultaneously deployed on a network of learning agents, each of which is designed to be capable of independently building local model from self-collected data and steadily improving its analytic quality through exchanging its model with other devices in the network. Finally, we showcase our empirical results via an assortment of practical scenarios, featuring both synthetic and real-world domains, which highlight the efficiency, resiliency and fault-tolerance of our framework. 

{\bf \noindent Acknowledgements.} This research is funded in part by ONR under BRC award $\#$N000141712072.

\bibliographystyle{natbib}
\bibliography{nips2018}

\begin{thebibliography}{}

\bibitem[Allamraju and Chowdhary(2017)Allamraju and Chowdhary]{Rakshit17}
Allamraju, R. and Chowdhary, G. (2017).
\newblock Communication efficient decentralized {Gaussian} process fusion for
  multi-{UAS} path planning.
\newblock In {\em Proc. {ACC}\/}.

\bibitem[Bui {\em et~al.\/}(2017)Bui, Nguyen, and Turner]{Bui17}
Bui, T.~D., Nguyen, C.~V., and Turner, R.~E. (2017).
\newblock Streaming sparse gaussian process approximations.
\newblock In {\em Proc. {NIPS}\/}.

\bibitem[Cao {\em et~al.\/}(2013)Cao, Low, and Dolan]{LowAAMAS13}
Cao, N., Low, K.~H., and Dolan, J.~M. (2013).
\newblock Multi-robot informative path planning for active sensing of
  environmental phenomena: A tale of two algorithms.
\newblock In {\em Proc. {AAMAS}\/}, pages 7--14.

\bibitem[Chen {\em et~al.\/}(2012)Chen, Low, Tan, Oran, Jaillet, Dolan, and
  Sukhatme]{LowUAI12}
Chen, J., Low, K.~H., Tan, C. K.-Y., Oran, A., Jaillet, P., Dolan, J.~M., and
  Sukhatme, G.~S. (2012).
\newblock Decentralized data fusion and active sensing with mobile sensors for
  modeling and predicting spatiotemporal traffic phenomena.
\newblock In {\em Proc. UAI\/}, pages 163--173.

\bibitem[Chen {\em et~al.\/}(2013a)Chen, Cao, Low, Ouyang, Tan, and
  Jaillet]{LowUAI13}
Chen, J., Cao, N., Low, K.~H., Ouyang, R., Tan, C. K.-Y., and Jaillet, P.
  (2013a).
\newblock Parallel {Gaussian} process regression with low-rank covariance
  matrix approximations.
\newblock In {\em Proc. UAI\/}, pages 152--161.

\bibitem[Chen {\em et~al.\/}(2013b)Chen, Low, and Tan]{LowRSS13}
Chen, J., Low, K.~H., and Tan, C. K.-Y. (2013b).
\newblock {Gaussian} process-based decentralized data fusion and active sensing
  for mobility-on-demand system.
\newblock In {\em Proc. {RSS}\/}.

\bibitem[Chen {\em et~al.\/}(2015)Chen, Low, Jaillet, and Yao]{Arik15}
Chen, J., Low, K.~H., Jaillet, P., and Yao, Y. (2015).
\newblock {Gaussian} process decentralized data fusion and active sensing for
  spatiotemporal traffic modeling and prediction in mobility-on-demand systems.
\newblock {\em IEEE Transactions on Automation Science and Engineering\/}, {\bf
  12}(3), 901--921.

\bibitem[{Csat\'{o}} and Opper(2002){Csat\'{o}} and Opper]{Opper02}
{Csat\'{o}}, L. and Opper, M. (2002).
\newblock Sparse online gaussian processes.
\newblock {\em Neural Computation\/}, {\bf 14}(3), 641--669.

\bibitem[Deisenroth and Ng(2015)Deisenroth and Ng]{Marc15}
Deisenroth, M.~P. and Ng, J.~W. (2015).
\newblock Distributed {Gaussian} processes.
\newblock In {\em Proc. {ICML}\/}.

\bibitem[Gal {\em et~al.\/}(2014)Gal, {van der Wilk}, and Rasmussen]{Yarin14}
Gal, Y., {van der Wilk}, M., and Rasmussen, C. (2014).
\newblock Distributed variational inference in sparse {G}aussian process
  regression and latent variable models.
\newblock In {\em Proc. NIPS\/}, pages 3257--3265.

\bibitem[Hensman {\em et~al.\/}(2013)Hensman, Fusi, and Lawrence]{Hensman13}
Hensman, J., Fusi, N., and Lawrence, N.~D. (2013).
\newblock Gaussian processes for big data.
\newblock In {\em Proc. UAI\/}, pages 282--290.

\bibitem[Hoang {\em et~al.\/}(2017)Hoang, Hoang, and Low]{NghiaAAAI17}
Hoang, Q.~M., Hoang, T.~N., and Low, K.~H. (2017).
\newblock A generalized stochastic variational {B}ayesian hyperparameter
  learning framework for sparse spectrum {G}aussian process regression.
\newblock In {\em Proc. {AAAI}\/}, pages 2007--2014.

\bibitem[Hoang {\em et~al.\/}(2014)Hoang, Low, Jaillet, and
  Kankanhalli]{NghiaICML14}
Hoang, T.~N., Low, K.~H., Jaillet, P., and Kankanhalli, M. (2014).
\newblock Nonmyopic $\epsilon$-{B}ayes-{O}ptimal {A}ctive {L}earning of
  {G}aussian {P}rocesses.
\newblock In {\em Proc. ICML\/}, pages 739--747.

\bibitem[Hoang {\em et~al.\/}(2015)Hoang, Hoang, and Low]{NghiaICML15}
Hoang, T.~N., Hoang, Q.~M., and Low, K.~H. (2015).
\newblock A unifying framework of anytime sparse {Gaussian} process regression
  models with stochastic variational inference for big data.
\newblock In {\em Proc. {ICML}\/}, pages 569--578.

\bibitem[Hoang {\em et~al.\/}(2016)Hoang, Hoang, and Low]{NghiaICML16}
Hoang, T.~N., Hoang, Q.~M., and Low, K.~H. (2016).
\newblock A distributed variational inference framework for unifying parallel
  sparse {G}aussian process regression models.
\newblock In {\em Proc. {ICML}\/}, pages 382--391.

\bibitem[Kang and Larkin(2016)Kang and Larkin]{Kang16}
Kang, J.~J. and Larkin, H. (2016).
\newblock Inference of personal sensors in internet of things.
\newblock {\em International Journal of Information, Communication Technology
  and Applications\/}, {\bf 2}, 1.

\bibitem[{L\'{a}zaro}-Gredilla {\em et~al.\/}(2010){L\'{a}zaro}-Gredilla,
  {Qui\~{n}onero}-Candela, Rasmussen, and Figueiras-Vidal]{Miguel10}
{L\'{a}zaro}-Gredilla, M., {Qui\~{n}onero}-Candela, J., Rasmussen, C.~E., and
  Figueiras-Vidal, A.~R. (2010).
\newblock Sparse spectrum {G}aussian process regression.
\newblock {\em Journal of Machine Learning Research\/}, pages 1865--1881.

\bibitem[Low {\em et~al.\/}(2012)Low, Chen, Dolan, Chien, and
  Thompson]{LowAAMAS12}
Low, K.~H., Chen, J., Dolan, J.~M., Chien, S., and Thompson, D.~R. (2012).
\newblock Decentralized active robotic exploration and mapping for
  probabilistic field classification in environmental sensing.
\newblock In {\em Proc. {AAMAS}\/}, pages 105--112.

\bibitem[Low {\em et~al.\/}(2015)Low, Yu, Chen, and Jaillet]{LowAAAI15}
Low, K.~H., Yu, J., Chen, J., and Jaillet, P. (2015).
\newblock Parallel {Gaussian} process regression for big data: Low-rank
  representation meets {M}arkov approximation.
\newblock In {\em Proc. {AAAI}\/}, pages 2821--2827.

\bibitem[Min and Wynter(2011)Min and Wynter]{min11}
Min, W. and Wynter, L. (2011).
\newblock Real-time road traffic prediction with spatio-temporal correlations.
\newblock {\em Transport. Res. C-Emer.\/}, {\bf 19}(4), 606--616.

\bibitem[{Qui\~{n}onero}-Candela and Rasmussen(2005){Qui\~{n}onero}-Candela and
  Rasmussen]{Candela05}
{Qui\~{n}onero}-Candela, J. and Rasmussen, C.~E. (2005).
\newblock A unifying view of sparse approximate {Gaussian} process regression.
\newblock {\em Journal of Machine Learning Research\/}, {\bf 6}, 1939--1959.

\bibitem[Rasmussen and Williams(2006)Rasmussen and Williams]{Rasmussen06}
Rasmussen, C.~E. and Williams, C. K.~I. (2006).
\newblock {\em Gaussian Processes for Machine Learning\/}.
\newblock MIT Press.

\bibitem[Robbins and Monro(1951)Robbins and Monro]{Monro1951}
Robbins, H. and Monro, S. (1951).
\newblock A stochastic approximation method.
\newblock In {\em The Annals of Mathematical Statistics\/}, pages 400--407.

\bibitem[Sarkar {\em et~al.\/}(2014)Sarkar, Nambi, Prasad, and Rahim]{Sarkar14}
Sarkar, C., Nambi, S. N. A.~U., Prasad, R.~V., and Rahim, A. (2014).
\newblock A scalable distributed architecture towards unifying {IoT}
  applications.
\newblock In {\em Proc. 2014 IEEE World Forum on Internet of Things
  (WF-IoT)\/}.

\bibitem[Snelson and Ghahramani(2007)Snelson and Ghahramani]{Snelson07a}
Snelson, E.~L. and Ghahramani, Z. (2007).
\newblock Local and global sparse {G}aussian process approximation.
\newblock In {\em Proc. AISTATS\/}.

\bibitem[Titsias(2009)Titsias]{Titsias09}
Titsias, M.~K. (2009).
\newblock Variational learning of inducing variables in sparse {G}aussian
  processes.
\newblock In {\em Proc. {AISTATS}\/}, pages 567--574.

\bibitem[Titsias and {L\'{a}zaro}-Gredilla(2013)Titsias and
  {L\'{a}zaro}-Gredilla]{Titsias13}
Titsias, M.~K. and {L\'{a}zaro}-Gredilla, M. (2013).
\newblock Variational inference for {M}ahalanobis distance metrics in
  {G}aussian process regression.
\newblock In {\em Proc. {NIPS}\/}.

\bibitem[Wang and Papageorgiou(2005)Wang and Papageorgiou]{wang05}
Wang, Y. and Papageorgiou, M. (2005).
\newblock Real-time freeway traffic state estimation based on extended {Kalman}
  filter: a general approach.
\newblock {\em Transport. Res. B-Meth.\/}, {\bf 39}(2), 141--167.

\bibitem[Work {\em et~al.\/}(2010)Work, Blandin, Tossavainen, Piccoli, and
  Bayen]{Bayen10a}
Work, D.~B., Blandin, S., Tossavainen, O., Piccoli, B., and Bayen, A. (2010).
\newblock A traffic model for velocity data assimilation.
\newblock {\em AMRX\/}, {\bf 2010}(1), 1--35.

\end{thebibliography}
	
\ifthenelse{\value{sol}=1}{

\clearpage		
\appendix
		
\section{Derivation of Eq.~\eqref{eq:3.5}}
\label{app:a}
By definition of $\mathbf{R}_1$ and the expression of $\mathbf{S}$ in Eq.~\eqref{eq:2.4}, we have:
\begin{eqnarray}
\hspace{-4mm}\mathbf{R}_1 \ \ \triangleq\ \ \mathbf{S}^{-1} &=& \frac{1}{\sigma_n^2}\mathbf{K}_\mathcal{UU}^{-1}(\sigma^2_n\mathbf{K}_\mathcal{UU} + \mathbf{C}_\mathcal{UU})\mathbf{K}_\mathcal{UU}^{-1} \ \ =\ \ \mathbf{K}_\mathcal{UU}^{-1} + \frac{1}{\sigma^2_n}\mathbf{K}_\mathcal{UU}^{-1}\mathbf{C}_\mathcal{UU}\mathbf{K}_\mathcal{UU}^{-1} \ .
\label{eq:a.19}
\end{eqnarray}
On the other hand, by definition, we also have:
\begin{eqnarray}
\mathbf{C}_\mathcal{UU} \hspace{-2mm}&=&\hspace{-2mm} \mathbb{E}_{\mathrm{q}(\mathbf{W})}\left[\mathbf{K}_\mathcal{DU}\mathbf{K}_\mathcal{UD}\right] = \mathbb{E}_{\mathrm{q}(\mathbf{W})}\left[\sum_{i=1}^p\mathbf{K}_{\mathcal{UD}_i}\mathbf{K}_{\mathcal{D}_i\mathcal{U}}\right] \nonumber\\
\hspace{-2mm}&=&\hspace{-2mm} \sum_{i=1}^p \mathbb{E}_{\mathrm{q}(\mathbf{W})}\left[\mathbf{K}_{\mathcal{UD}_i}\mathbf{K}_{\mathcal{D}_i\mathcal{U}}\right] = \sum_{i=1}^p \mathbf{C}^i_\mathcal{UU} \ .
\label{eq:a.20}
\end{eqnarray}
Plugging Eq.~\eqref{eq:a.20} into Eq.~\eqref{eq:a.19} yields
\begin{eqnarray}
\mathbf{R}_1 &=& \mathbf{K}^{-1}_\mathcal{UU} + \frac{1}{\sigma_n^2} \sum_{i=1}^p \mathbf{K}^{-1}_\mathcal{UU}\mathbf{C}^{i}_\mathcal{UU}\mathbf{K}^{-1}_\mathcal{UU} \ .
\end{eqnarray}
By definition of $\mathbf{R}_2$ and the expression of $\mathbf{S}$ and $\mathbf{m}$ in Eq.~\eqref{eq:2.4}, we have:
\begin{eqnarray}
\mathbf{R}_2 \ \ \triangleq\ \ \mathbf{S}^{-1}\mathbf{m} &=& \frac{1}{\sigma_n^2}\mathbf{K}^{-1}_\mathcal{UU}\mathbf{C}_\mathcal{UD}\mathbf{y}_\mathcal{D} \ .
\label{eq:a.22}
\end{eqnarray}
Again, by definition, we also have:
\begin{eqnarray}
\hspace{-4mm}\mathbf{C}_\mathcal{UD}\mathbf{y}_\mathcal{D} &=& \mathbb{E}_{q(\mathbf{W})}\left[\sum_{i=1}^p \mathbf{K}_{\mathcal{UD}_i}\mathbf{y}_{\mathcal{D}_i}\right] \ =\ \sum_{i=1}^p \mathbb{E}_{\mathrm{q}(\mathbf{W})}\left[\mathbf{K}_{\mathcal{UD}_i}\right]\mathbf{y}_{\mathcal{D}_i}  \ =\ \sum_{i=1}^p \mathbf{C}_{\mathcal{UD}_i}\mathbf{y}_{\mathcal{D}_i} \ .
\label{eq:a.23}
\end{eqnarray}
Plugging Eq.~\eqref{eq:a.23} into Eq.~\eqref{eq:a.22}, we have
\begin{eqnarray}
\mathbf{R}_2 &=& \frac{1}{\sigma^2_n}\sum_{i=1}^p\mathbf{K}^{-1}_\mathcal{UU}\mathbf{C}_{\mathcal{UD}_i}\mathbf{y}_{\mathcal{D}_i} \ ,
\end{eqnarray}
which concludes our derivation.

\section{Derivation of $\mathrm{L}(q)$'s decomposability}
\label{app:b}
By definition, we have 
\begin{eqnarray}
\mathrm{L}(q) &=& \mathbb{E}_\mathrm{q} \left[ \mathrm{log} \  \mathrm{p}(\mathbf{y}_\mathcal{D}| \mathbf{f}_\mathcal{D})\right] - \mathrm{D}_{\mathrm{KL}} (\mathrm{q}(\mathbf{u,W}) \| \mathrm{p}(\mathbf{u,W}))
\label{eq:b.25}
\end{eqnarray} 
where the expectation is with respect to $\mathrm{q} \triangleq \mathrm{q}(\mathbf{f}_\mathcal{D},\mathbf{u,W}) \triangleq \mathrm{q}(\mathbf{u,W})\mathrm{p}(\mathbf{f}_\mathcal{D} | \mathbf{u,W})$. The first term on the RHS of Eq.~\eqref{eq:b.25} can be rewritten more concisely as $\mathbb{E}_{\mathrm{q}} \left[ \mathrm{log} \ \mathrm{p}({\mathbf{y}_\mathcal{D} | \mathbf{f}_\mathcal{D}}) \right] = $
\begin{eqnarray}
\hspace{-6mm}\mathbb{E}_{\mathrm{q}} \left[ \mathrm{log} \ \mathrm{p}({\mathbf{y}_\mathcal{D} | \mathbf{f}_\mathcal{D}}) \right] &=& \mathbb{E}_{\mathrm{q}(\mathbf{u, W})}\mathbb{E}_{\mathrm{p}(\mathbf{f}_\mathcal{D} | \mathbf{u, W})} \left[ \sum_{i=1}^N \mathrm{log}\ \mathrm{p} (\mathbf{y}_{\mathcal{D}_i} | \mathbf{f}_{\mathcal{D}_i}) \right] \nonumber \\
&=& \mathbb{E}_{\mathrm{q}(\mathbf{u, W})} \left[\sum_{i=1}^N \mathbb{E}_{\mathrm{p}(\mathbf{f}_\mathcal{D} | \mathbf{u, W})} \left[ \mathrm{log} \ \mathrm{p} (\mathbf{y}_{\mathcal{D}_i} | \mathbf{f}_{\mathcal{D}_i}) \right]\right] \nonumber \\ 
&=& \sum_{i=1}^N\mathbb{E}_{\mathrm{q}(\mathbf{u, W})} \left[ \mathbb{E}_{\mathrm{p}(\mathbf{f}_{\mathcal{D}_i} | \mathbf{u, W})} \left[ \mathrm{log} \ \mathrm{p} (\mathbf{y}_{\mathcal{D}_i} | \mathbf{f}_{\mathcal{D}_i}) \right]\right] \ =\ \sum_{i=1}^N \mathrm{L}_{\mathcal{D}_i}(\mathrm{q}) \ ,
\label{eq:b.26}
\end{eqnarray}
where the second last equality follows from the fact that given $\mathbf{u}$ and $\mathbf{W}$, $\mathbf{f}_{\mathcal{D}_i} \perp \mathbf{f}_{\mathcal{D}_j} \forall \ {i \neq j}$ and the last equality follows directly from the definition of $\mathrm{L}_{\mathcal{D}_i}(\mathrm{q})$. Finally, plugging Eq.~\eqref{eq:b.26} into Eq.~\eqref{eq:b.25} yields the desired result.

\section{Proof of $\mathbb{E}_{\mathcal{D}_\ast}[\partial\widehat{\mathrm{L}}(\mathrm{q})/\partial\theta] = \partial\mathrm{L}(\mathrm{q})/\partial\theta$}
\label{app:c}
Since $\mathcal{D}_\ast$ is sampled uniformly from $\{\mathcal{D}_1,\mathcal{D}_2, \ldots, \mathcal{D}_N\}$, we have $\mathrm{Pr}\left(\mathcal{D}_\ast = \mathcal{D}_i \right) = 1 / N$. Hence, 
\begin{eqnarray}
\hspace{-4mm}\mathbb{E}_{\mathcal{D}_\ast}[\partial\widehat{\mathrm{L}}(\mathrm{q})/\partial\theta] &=& \frac{1}{N}\sum_{i=1}^N \left(N\frac{\partial\mathrm{L}_{\mathcal{D}_i}(\mathrm{q})}{\partial\theta} - \frac{\partial}{\partial\theta}\mathrm{D}_{\mathrm{KL}}\left(\mathrm{q}(\mathbf{u,W}) \| \mathrm{p}(\mathbf{u, W} \right)\right) \nonumber \\
&=& - \frac{\partial}{\partial\theta}\mathrm{D}_{\mathrm{KL}}\left(\mathrm{q}(\mathbf{u,W}) \| \mathrm{p}(\mathbf{u, W} \right) + \sum_{i=1}^N \frac{\partial\mathrm{L}_{\mathcal{D}_i}(\mathrm{q})}{\partial\theta} \ \ =\ \ \frac{\partial\mathrm{L}(\mathrm{q})}{\partial\theta} \ ,
\end{eqnarray}
which completes our proof.

\section{Derivation of Pairwise Fusion Formula} 
\label{app:d}
Applying Bayes Theorem, we have :
\begin{eqnarray}
\mathrm{p}(\mathbf{u}|\mathbf{y}_{\mathcal{D}_a}, \mathbf{y}_{\mathcal{D}_b}) \hspace{-1mm}&=&\hspace{-1mm} \frac{\mathrm{p}(\mathbf{y}_{\mathcal{D}_a}, \mathbf{y}_{\mathcal{D}_b}|\mathbf{u})p(\mathbf{u})}{\mathrm{p}(\mathbf{y}_{\mathcal{D}_a}, \mathbf{y}_{\mathcal{D}_b})} \ =\ \frac{\mathrm{p}(\mathbf{y}_{\mathcal{D}_a} | \mathbf{u}) \mathrm{p}(\mathbf{y}_{\mathcal{D}_b}|\mathbf{u})\mathrm{p}(\mathbf{u})}{\mathrm{p}(\mathbf{y}_{\mathcal{D}_a}, \mathbf{y}_{\mathcal{D}_b})} \nonumber \\
\hspace{-1mm}&=&\hspace{-1mm} 
\frac{\mathrm{p}(\mathbf{u}|\mathbf{y}_{\mathcal{D}_a}) \mathrm{p}(\mathbf{y}_{\mathcal{D}_a})\mathrm{p}(\mathbf{u}|\mathbf{y}_{\mathcal{D}_b}) \mathrm{p}(\mathbf{y}_{\mathcal{D}_b})\mathrm{p}(\mathbf{u})}{\mathrm{p}(\mathbf{u})^2\mathrm{p}(\mathbf{y}_{\mathcal{D}_a},\mathbf{y}_{\mathcal{D}_b})} \ \propto\ \frac{\mathrm{p}(\mathbf{u}|\mathbf{y}_{\mathcal{D}_a}) \mathrm{p}(\mathbf{u}|\mathbf{y}_{\mathcal{D}_b}) }{\mathrm{p}(\mathbf{u})} \ ,
\end{eqnarray}
which completes our derivation.
\section{Derivation of Eq.~\eqref{eq:4.10}}
\label{app:e}
By definition, $\mathrm{q}_{ab}(\mathbf{u}) \propto \mathrm{q}_{a}(\mathbf{u})\mathrm{q}_{b}(\mathbf{u})/\mathrm{p}(\mathbf{u})$ where we have the approximate posteriors $\mathrm{q}_a(\mathbf{u}) = \mathcal{N}(\mathbf{u};\mathbf{m}_a, \mathbf{S}_a)$, $\mathrm{q}_b(\mathbf{u}) = \mathcal{N}(\mathbf{u};\mathbf{m}_b, \mathbf{S}_b)$ and prior $\mathrm{p}(\mathbf{u}) = \mathcal{N}(\mathbf{u};0, \mathbf{K}_\mathcal{UU})$. 

As such, we have $\log \mathrm{q}_{ab}(\mathbf{u}) \propto \mathrm{log}(\mathrm{q}_a(\mathbf{u})\mathrm{q}_b(\mathbf{u})/\mathrm{p}(\mathbf{u})) \propto$
\begin{eqnarray} \hspace{-0.5mm}-\frac{1}{2}\mathbf{u}^\top(\mathbf{S}_a^{-1} + \mathbf{S}_b^{-1} - \mathbf{K}^{-1}_\mathcal{UU})\mathbf{u} + \mathbf{u}^\top(\mathbf{S}_a^{-1}\mathbf{m}_a + \mathbf{S}_b^{-1}\mathbf{m}_b) \label{eq:e.1}
\end{eqnarray}
On the other hand, we also have
\begin{eqnarray}
\hspace{-2mm}\log \mathrm{q}_{ab}(\mathbf{u}) &\propto& -\frac{1}{2}\mathbf{u}^\top\mathbf{S}_{ab}^{-1}\mathbf{u} + \mathbf{u}^\top\mathbf{S}_{ab}^{-1}\mathbf{m}_{ab} \label{eq:e.2}
\end{eqnarray}
Matching Eq.~\eqref{eq:e.1} with Eq.~\eqref{eq:e.2}, we have $\mathrm{q}_{ab}(\mathbf{u}) = \mathcal{N}(\mathbf{u}; \mathbf{m}_{ab}, \mathbf{S}_{ab})$, where:
\begin{eqnarray}
\mathbf{S}_{ab} &=& (\mathbf{S}_a^{-1} + \mathbf{S}_b^{-1} - \mathbf{K}^{-1}_\mathcal{UU})^{-1} \ , \nonumber \\
\mathbf{m}_{ab} &=& \mathbf{S}_{ab} (\mathbf{S}^{-1}_a\mathbf{m}_a + \mathbf{S}^{-1}_b\mathbf{m}_b) \ .
\end{eqnarray}
This completes our derivation.
\section{Derivation of Multi-Agent Fusion Formula}
\label{app:f}
It is straight-forward to see that $\mathbf{R}_g = \sum_{i=1}^s \mathbf{R}_i - (s-1)\mathbf{R}_0$ is true for $s=2$ by applying the pair-wise fusion formula in Section~\ref{pairwise}. Suppose this is also true for $s=k$, we proceed to prove by induction that it is also true for $s=k+1$. That is, given $k$ agents whose natural representations are $\mathbf{R}_1,\mathbf{R}_2,\ldots,\mathbf{R}_{k}$ respectively and let $\mathbf{R}_a$ be the natural representation of their fused model $\mathrm{q}_a(\mathbf{u}) \simeq \mathrm{p}(\mathbf{u}|\mathbf{y}_{\mathcal{D}_1},\mathbf{y}_{\mathcal{D}_2},\ldots,\mathbf{y}_{\mathcal{D}_k}) = \mathrm{p}(\mathbf{u}|\mathbf{y}_{\mathcal{D}_a})$ with $\mathcal{D}_a \triangleq \{\mathcal{D}_i\}_{i=1}^k$. 

Applying our inductive assumption for $s = k$:
\begin{eqnarray}
\mathbf{R}_a &=& \left(\sum_{i=1}^k \mathbf{R}_i \right) - (k - 1)\mathbf{R}_0
\label{eq:f.31}
\end{eqnarray}
Then, let us denote $\mathcal{D}_b \triangleq \mathcal{D}_{k+1}$ and note that $\mathrm{p}(\mathbf{u} | \mathbf{y}_{\mathcal{D}_1}, \mathbf{y}_{\mathcal{D}_2}, \ldots, \mathbf{y}_{\mathcal{D}_{k+1}}) \propto \mathrm{p}(\mathbf{u}|\mathbf{y}_{\mathcal{D}_a})\ \mathrm{p}(\mathbf{u}|\mathbf{y}_{\mathcal{D}_b}) / \mathrm{p}(\mathbf{u})$, which is approximated by $\mathrm{q}_{ab}(\mathbf{u}) \propto \mathrm{q}_a(\mathbf{u})\mathrm{q}_b(\mathbf{u}) / \mathrm{p}(\mathbf{u})$. Thus, let $\mathbf{R}_{ab}$ denote the natural representation of $\mathrm{q}_ab(\mathbf{u})$, we have $\mathbf{R}_g = \mathbf{R}_{ab}$. Then, let $\mathbf{R}_b$ denote the natural representation of the $\mathrm{q}_{b}(\mathbf{u})$, we have
\begin{eqnarray}
\mathbf{R}_g \ \ =\ \ \mathbf{R}_{ab} &=& \mathbf{R}_a + \mathbf{R}_b - \mathbf{R}_0 \ .
\label{eq:f.32}
\end{eqnarray}
Plugging Eq.~\eqref{eq:f.31} into Eq.~\eqref{eq:f.32} finally yields
\begin{eqnarray}
\mathbf{R}_g &=& \left(\sum_{i=1}^k \mathbf{R}_i \right) - (k - 1)\mathbf{R}_0 + \mathbf{R}_{k+1} - \mathbf{R}_0 \nonumber \\
&=& \left(\sum_{i=1}^{k+1} \mathbf{R}_i \right) - k\mathbf{R}_0 \ ,
\end{eqnarray}
which proves that the result also holds for $s = k + 1$. By induction, this means it will hold for all $s$. 

\section{Proof of Lemma~\ref{lem1}}
\label{app:g}
By definition of $\widehat{\mathbf{C}}^i_\mathcal{UU}$, we have:
\begin{eqnarray}
\hspace{-8mm}\mathbb{E}_\mathrm{p}(\mathbf{W})\left[\widehat{\mathbf{C}}^i_\mathcal{UU} \right] &=& \frac{1}{k} \sum_{t=1}^k \mathbb{E}_{\mathrm{p}(\mathbf{W})}\left[\frac{\mathrm{q}(\mathbf{W}_t)}{\mathrm{p}(\mathbf{W}_t)}\mathbf{K}^{(t)}_{\mathcal{UD}_i}\mathbf{K}^{(t)}_{\mathcal{D}_i\mathcal{U}} \right] \ =\ \mathbb{E}_{\mathrm{p}(\mathbf{W})} \left[\frac{\mathrm{q}(\mathbf{W})}{\mathrm{p}(\mathbf{W})} \mathbf{K}_{\mathcal{UD}_i}\mathbf{K}_{\mathcal{D}_i\mathcal{U}}\right] \nonumber \\
&=& \mathbb{E}_{\mathrm{q}(\mathbf{W})}\left[ \mathbf{K}_{\mathcal{UD}_i}\mathbf{K}_{\mathcal{D}_i\mathcal{U}} \right] \ \triangleq\ \mathbf{C}^i_\mathcal{UU}
\end{eqnarray}
where the second equality follows from the fact that $\{\mathbf{W}_t\}_{t}$ are identically and independently drawn from $\mathrm{p}(\mathbf{W})$. On the other hand, let $\mathbf{R} \triangleq [\mathbf{R}_1; \mathbf{R}_2]$ denote the local representation of an arbitrary agent, we have
\begin{eqnarray}
\hspace{-2.5mm}\mathbb{E}_{\mathrm{p}(\mathbf{W})}\left[\widehat{\mathbf{R}}_1\right] &=& \mathbf{K}^{-1}_\mathcal{UU} + \sum_{i=1}^p \frac{1}{\sigma_n^2} \mathbf{K}^{-1}_\mathcal{UU}\mathbb{E}_{\mathrm{p}(\mathbf{W})}\left[\widehat{\mathbf{C}}^i_\mathcal{UU}\right]\mathbf{K}^{-1}_\mathcal{UU} \nonumber \\
&=& \mathbf{K}^{-1}_{\mathcal{UU}} + \sum_{i=1}^p\frac{1}{\sigma_n^2} \mathbf{K}^{-1}_\mathcal{UU} \mathbf{C}^i_\mathcal{UU}\mathbf{K}^{-1}_\mathcal{UU} \triangleq \mathbf{R}_1
\end{eqnarray}
Using similar reasoning, we also have $\mathbb{E}_{\mathrm{p}(\mathbf{W})}[\widehat{\mathbf{R}}_2] = \mathbf{R}_2$. It immediately implies that
$\mathbb{E}_{\mathrm{p}(\mathbf{W})}[\widehat{\mathbf{R}}] = \mathbb{E}_{\mathrm{p}(\mathbf{W})}[\widehat{\mathbf{R}}_1; \widehat{\mathbf{R}}_2] = [\mathbf{R}_1; \mathbf{R}_2] \triangleq \mathbf{R}$. This also implies, for any vector component $\mathbf{R}(i)$ and $\widehat{\mathbf{R}}(i)$ of $\mathbf{R}, \widehat{\mathbf{R}}$ (assuming $\mathbf{R}$ and $ \widehat{\mathbf{R}}$ are vectorized), we have $\mathbb{E}[\widehat{\mathbf{R}}(i)] = \mathbf{R}(i)$ where $1 \leq i \leq |\mathbf{R}| = |\mathbf{R}_1| + |\mathbf{R}_2| = m(m+1)$. Applying Hoeffding inequality for each vector component $\mathbf{R}(i)$ and its unbiased estimation $\widehat{\mathbf{R}}(i)$, we have:
\begin{eqnarray}
\hspace{-1.5mm}\mathrm{Pr}\left(\left|\mathbf{R}(i) - \widehat{\mathbf{R}}(i)\right| \leq \epsilon'\right) &\geq& 1 - 2\mathrm{exp}\left(-\frac{2k{\epsilon'}^2}{C} \right) \ ,
\label{eq:g.36}
\end{eqnarray}
assuming $\widehat{\mathbf{R}}(i)$ is bounded above and below and the size of the bounding interval is upper-bounded by a sufficiently large constant $C > 0$. Let choose $\delta \in (0, 1)$ for which $\delta/(m(m+1)) = 2\mathrm{exp}(-2k\epsilon'^2/C)$. Then, it follows that, for each vector index $i \in [1, m(m+1)]$, by choosing $k = (1/\epsilon'^2)\mathrm{log} (2 m(m+1)/\delta)  = \mathcal{O}((1/\epsilon')^2\mathrm{log}(m/\delta))$, the inequality $|\mathbf{R}(i) - \widehat{\mathbf{R}}(i)| \leq \epsilon'$ holds with probability at least $1 - \delta/(m(m+1))$. Then, by union bound, $|\mathbf{R}(i) - \widehat{\mathbf{R}}(i)| \leq \epsilon'$ holds simultaneously for all $i$ with probability at least $1 - \delta$. When that happens, we have:
\begin{eqnarray}
\| \mathbf{R} - \widehat{\mathbf{R}} \|^2 &=& \sum_{i=1}^{m(m+1)} |\mathbf{R}(i) - \widehat{\mathbf{R}}(i)|^2\ \ \leq\ \ m(m+1) {\epsilon'}^2 
\end{eqnarray}
Finally, let $\epsilon = \epsilon' * \sqrt{m(m+1)}$, we have:
\begin{eqnarray}
\mathrm{Pr}(\|\mathbf{R} - \widehat{\mathbf{R}}\| \leq \epsilon) &\leq& 1 - \delta
\end{eqnarray}
when $k = \mathcal{O}((m/\epsilon)^2\mathrm{log}(m/\delta))$. Setting $\mathbf{R} = \mathbf{R}_i$ for each agent $i$ thus concludes our proof.

\section{Proof of Theorem~\ref{theo1}}
\label{app:h}
We have $\widehat{\mathbf{R}}_g = \sum_{i=1}^s\widehat{\mathbf{R}}_i - (s-1)\mathbf{R}_0$ and
$\mathbf{R}_g = \sum_{i=1}^s\mathbf{R}_i - (s-1)\mathbf{R}_0$ which immediately implies
\begin{eqnarray}
\|\mathbf{R}_g - \widehat{\mathbf{R}}_g\| &\leq& \sum_{i=1}^s \|\mathbf{R}_i - \widehat{\mathbf{R}}_i\| \ .
\end{eqnarray}
For each local representation $\mathbf{R}_i$, applying Lemma~\ref{lem1} with $\epsilon/s$ and $\delta/s$, we have:
\begin{eqnarray}
\mathrm{Pr}\left(\|\mathbf{R}_i - \widehat{\mathbf{R}}_i\| \leq \frac{\epsilon}{s}\right) &\geq& 1 - \frac{\delta}{s} \ ,
\end{eqnarray}
with $k = \mathcal{O}((ms/\epsilon)^2\mathrm{log}(ms/\epsilon))$. Then, applying union bound over the entire set of local representation $\{\mathbf{R}_i\}_{i=1}^s$, we have $\|\mathbf{R}_i - \widehat{\mathbf{R}}_i\| \leq \epsilon/s$ holds simultaneously for all $i$ with probability at least $1-\delta$. When that happens, we have
\begin{eqnarray}
\hspace{-10mm}\|\mathbf{R}_g - \widehat{\mathbf{R}}_g\| &\leq& \sum_{i=1}^s \|\mathbf{R}_i - \widehat{\mathbf{R}}_i\| \ \ \leq\ \ s\frac{\epsilon}{s} \ \ =\ \ \epsilon \ .
\end{eqnarray}
Thus, by choosing $k = \mathcal{O}((ms/\epsilon)^2\mathrm{log}(ms/\epsilon))$, we have
\begin{eqnarray}
\mathrm{Pr}\left(\|\mathbf{R}_g - \widehat{\mathbf{R}}_g\| \leq \epsilon\right) &\geq& 1 - \delta \ ,
\end{eqnarray}
which concludes our proof.		
}{}
	
\end{document}